%% file: main.tex
\PassOptionsToPackage{dvipsnames,table}{xcolor}
\documentclass{applemlr}

\input{apple_preamble}
\input{content_preamble}

\title{Simulate to Generalize: Scaling Stateful Supervision for API-calling Agents using LLM World Models}

\author{Seanie Lee}
\author{Sanjoy Chowdhury}
\author{Chao Jiang}
\author{Cheng-Yu Hsieh}
\author{Ting-Yao Hu}
\author{Alexander Toshev}
\author{Oncel Tuzel}
\author{Raviteja Vemulapalli}

\affiliation{Apple}

\abstract{
Training agents that generalize to unseen, stateful environments requires a massive dataset of state-changing trajectories covering a vast and diverse set of APIs. However, scaling this broad supervision is severely bottlenecked by the immense effort required to implement and populate fully-executable environments across a broad spectrum of domains. To bypass this barrier, we introduce a data generation pipeline that decouples data synthesis from environment construction by leveraging LLMs as digital world models. Starting from only a list of broad domain names, our automated pipeline synthesizes diverse APIs and tasks. To produce trajectories, a teacher agent iteratively solves these tasks while an LLM simulator dynamically tracks state and provides coherent API responses on-the-fly. Finally, an automated judge filters the trajectories for quality. Fine-tuning on our broad synthetic dataset yields significant performance gains on AppWorld and OfficeBench, two challenging stateful benchmarks featuring environments completely unseen during training. These results establish our LLM world model-based synthesis approach as a highly scalable path for training generalizable, stateful API-calling agents.}

\metadata[Correspondence]{\sffamily Seanie Lee: \url{seanie_lee@apple.com}; Raviteja Vemulapalli: \url{r_vemulapalli@apple.com}}
\date{\sffamily\today}

\begin{document}

\maketitle
\bibliographystyle{acl_natbib}
\input{sections/1_intro}

\input{sections/3_related_work}
\input{sections/2_method}

\input{sections/4_experiment}
\input{sections/5_conclusion}

\bibliography{reference}
\appendix
\input{sections/6_appendix}

\end{document}

%% file: apple_preamble.tex
\usepackage{amsmath}
\usepackage{enumerate}
\usepackage{algorithm}
\usepackage{algpseudocode}
\usepackage{amsfonts}
\usepackage{amsthm}
\usepackage{cleveref}
\usepackage{diagbox}
\usepackage{colortbl}
\usepackage{amssymb}
\usepackage{xspace}
\usepackage{wrapfig}
\usepackage{adjustbox}
\usepackage{tabularx}
\usepackage{booktabs}
\usepackage{mathtools}
\usepackage{tikz}
\usepackage{enumitem}
\usepackage{silence}
\usepackage{dsfont}
\usepackage[table]{xcolor}
\usepackage[dvipsnames]{xcolor}
\usepackage{multirow}
\usepackage{makecell}
\usepackage{xfakebold}
\input{math_commands}

\definecolor{textgray}{HTML}{6E6E73}
\usetikzlibrary{positioning, calc}
\usetikzlibrary{decorations.pathmorphing}

\makeatletter
\patchcmd{\wrong@fontshape}{\@gobbletwo}{}{}{}
\makeatother
\numberwithin{equation}{section}
\makeatletter
\AtBeginDocument{
  \urlstyle{sf}
  
}
\makeatother

\definecolor{light}{RGB}{125, 125, 125}
\crefname{tcb@cnt@pbox}{code}{code}
\Crefname{tcb@cnt@pbox}{Code}{Code}
\crefname{assumption}{assumption}{assumption}
\Crefname{assumption}{Assumption}{Assumptions}

\newtcolorbox[auto counter]{pbox}[2][]{
  colback=white,
  title=Code~\thetcbcounter: #2,
  #1,fonttitle=\sffamily,
  fontupper=\sffamily,
  arc=2pt,
  colframe=bgcolor,
  coltitle=fgcolor,
  colbacktitle=bgcolor,
  toptitle=0.25cm,
  bottomtitle=0.125cm
}

\makeatletter
\newcommand\applefootnote[1]{%
  \begingroup
  \renewcommand\thefootnote{}%
  \renewcommand\@makefntext[1]{\noindent##1}%
  \footnote{#1}%
  \addtocounter{footnote}{-1}%
  \endgroup
}
\makeatother

\definecolor{cverbbg}{gray}{0.90}

\usetikzlibrary{arrows.meta, positioning, shapes.geometric, fit, backgrounds, shadows}
\usepackage{upquote}
\usepackage{subcaption}
\usepackage[dvipsnames]{xcolor} 
\usepackage{comment}
\usepackage{geometry}
\usepackage{placeins}
\definecolor{utilcolor}{rgb}{0.9,1.0,0.9}

\definecolor{Red}{rgb}{0.835, 0.180, 0.180}
\definecolor{Blue}{rgb}{0.227, 0.502, 0.824}
\definecolor{Green}{rgb}{0.345, 0.647, 0.208}

\definecolor{stage1}{RGB}{217,119,6}      
\definecolor{stage1bg}{RGB}{254,243,199}  
\definecolor{stage2}{RGB}{37,99,235}      
\definecolor{stage2bg}{RGB}{219,234,254}  
\definecolor{stage3}{RGB}{22,163,74}      
\definecolor{stage3bg}{RGB}{220,252,231}  
\definecolor{typecolor}{RGB}{13,148,136}  
\definecolor{stringcolor}{RGB}{180,83,9}  

\crefname{section}{Sec.}{Secs.}
\crefname{figure}{Fig.}{Figs.}
\crefname{table}{Table}{Tables}
\crefname{lstlisting}{listing}{listings}

\crefname{appendix}{Appendix}{Appendices}
\Crefname{appendix}{Appendix}{Appendices}
\crefname{subappendix}{Appendix}{Appendices}
\Crefname{subappendix}{Appendix}{Appendices}
\crefname{subsubappendix}{Appendix}{Appendices}
\Crefname{subsubappendix}{Appendix}{Appendices}

\usepackage{booktabs}
\definecolor{mypink}{HTML}{FDECEC}
\definecolor{mygray}{HTML}{F2F2F2}

\newcommand{\eg}{\textit{e.g.}}
\newcommand{\ie}{\textit{i.e.}}

\usepackage{fvextra}          
\usepackage{tcolorbox}
\usepackage{caption}          
\tcbuselibrary{breakable, skins}

\definecolor{agentbg}{RGB}{227, 238, 255}
\definecolor{agentframe}{RGB}{100, 140, 220}
\definecolor{outputbg}{RGB}{237, 247, 237}
\definecolor{outputframe}{RGB}{100, 180, 100}
\definecolor{taskbg}{RGB}{250, 245, 235}
\definecolor{taskframe}{RGB}{180, 140, 80}

\newtcolorbox{trajbox}[1]{
    enhanced, breakable,
    colback=taskbg, colframe=taskframe,
    fonttitle=\bfseries\small, title=#1,
    boxrule=0.5pt, arc=2pt,
    top=4pt, bottom=4pt, left=6pt, right=6pt,
    before skip=4pt, after skip=4pt,
}
\newtcolorbox{agentbox}[1]{
    enhanced, breakable,
    colback=agentbg, colframe=agentframe,
    fonttitle=\bfseries\small, title=#1,
    boxrule=0.4pt, arc=2pt,
    top=3pt, bottom=3pt, left=6pt, right=6pt,
    before skip=2pt, after skip=1pt,
}
\newtcolorbox{envbox}[1]{
    enhanced, breakable,
    colback=outputbg, colframe=outputframe,
    fonttitle=\bfseries\small, title=#1,
    boxrule=0.4pt, arc=2pt,
    top=3pt, bottom=3pt, left=6pt, right=6pt,
    before skip=1pt, after skip=2pt,
}

\usepackage{makecell}

\usepackage{longtable}
\usepackage{array}
\usepackage{listings}

\definecolor{jsonstring}{RGB}{0,120,0}
\definecolor{jsonnumber}{RGB}{180,90,0}
\definecolor{jsonbool}{RGB}{0,140,140}
\definecolor{codebg}{RGB}{250,250,250}
\definecolor{promptbg}{RGB}{245,245,255}
\lstdefinelanguage{json}{
    basicstyle=\ttfamily\footnotesize,
    backgroundcolor=\color{codebg},
    showstringspaces=false,
    breaklines=true,
    frame=single, framesep=4pt,
    rulecolor=\color{gray!40},
    morestring=[b]",
    stringstyle=\color{jsonstring},
    literate=
        *{0}{{{\color{jsonnumber}0}}}{1}
         {1}{{{\color{jsonnumber}1}}}{1}
         {2}{{{\color{jsonnumber}2}}}{1}
         {3}{{{\color{jsonnumber}3}}}{1}
         {4}{{{\color{jsonnumber}4}}}{1}
         {5}{{{\color{jsonnumber}5}}}{1}
         {6}{{{\color{jsonnumber}6}}}{1}
         {7}{{{\color{jsonnumber}7}}}{1}
         {8}{{{\color{jsonnumber}8}}}{1}
         {9}{{{\color{jsonnumber}9}}}{1}
         {true}{{{\color{jsonbool}true}}}{4}
         {false}{{{\color{jsonbool}false}}}{5}
         {null}{{{\color{gray}null}}}{4},
}
\lstdefinestyle{prompt}{
    basicstyle=\ttfamily\footnotesize,
    backgroundcolor=\color{promptbg},
    showstringspaces=false,
    breaklines=true,
    frame=single, framesep=4pt,
    rulecolor=\color{gray!40},
}

\definecolor{promptbg}{RGB}{249,250,251}
\definecolor{promptaccent}{RGB}{37,99,235}
\definecolor{promptcaption}{RGB}{30,58,138}

\lstdefinestyle{prettyprompt}{
    basicstyle=\ttfamily\footnotesize,
    backgroundcolor=\color{promptbg},
    showstringspaces=false,
    breaklines=true,
    breakatwhitespace=true,
    columns=fullflexible,
    keepspaces=true,
    frame=l,
    framerule=2.5pt,
    rulecolor=\color{promptaccent},
    framesep=10pt,
    xleftmargin=15pt,
    xrightmargin=4pt,
    aboveskip=14pt,
    belowskip=14pt,
    abovecaptionskip=2pt,
    belowcaptionskip=8pt,
    captionpos=t,
    extendedchars=true,
    inputencoding=utf8,
    literate={—}{{\textemdash}}1
            {–}{{\textendash}}1
            {’}{{'}}1,
}

\lstdefinestyle{prompt}{...}    

%% file: math_commands.tex
\usepackage{amsmath,amsfonts,bm}

\def\eqref#1{equation~\ref{#1}}

\def\1{\bm{1}}

\DeclareMathAlphabet{\mathsfit}{\encodingdefault}{\sfdefault}{m}{sl}
\SetMathAlphabet{\mathsfit}{bold}{\encodingdefault}{\sfdefault}{bx}{n}



%% file: content_preamble.tex
\definecolor{stage0}{RGB}{124,58,237}     
\definecolor{stage0bg}{RGB}{237,233,254}  

\newif\ifarxiv
\arxivtrue

\newcommand{\trajcodesize}{\footnotesize}
\newcommand{\trajtextsize}{\small}

\newcommand{\narrowtablewidth}{0.62\linewidth}

\newcommand{\twotables}[6]{%
  \begin{table}[t]
  \centering
  \begin{minipage}[t]{0.48\linewidth}
    \centering\small #1
    \caption{#2}\label{#3}
  \end{minipage}\hfill
  \begin{minipage}[t]{0.48\linewidth}
    \centering\small #4
    \caption{#5}\label{#6}
  \end{minipage}
  \end{table}%
}

\newenvironment{domainfig}{\figure[t]}{\endfigure}
\newcommand{\domainfigwidth}{0.6\linewidth}

\newtcolorbox{trajboxf}[1]{
    enhanced,
    colback=taskbg, colframe=taskframe,
    fonttitle=\bfseries\small, title=#1,
    boxrule=0.5pt, arc=2pt,
    top=4pt, bottom=4pt, left=6pt, right=6pt,
    before skip=4pt, after skip=4pt,
}
\newtcolorbox{agentboxf}[1]{
    enhanced,
    colback=agentbg, colframe=agentframe,
    fonttitle=\bfseries\small, title=#1,
    boxrule=0.4pt, arc=2pt,
    top=3pt, bottom=3pt, left=6pt, right=6pt,
    before skip=2pt, after skip=1pt,
}
\newtcolorbox{envboxf}[1]{
    enhanced,
    colback=outputbg, colframe=outputframe,
    fonttitle=\bfseries\small, title=#1,
    boxrule=0.4pt, arc=2pt,
    top=3pt, bottom=3pt, left=6pt, right=6pt,
    before skip=1pt, after skip=2pt,
}

%% file: sections/1_intro.tex
\ifarxiv
\input{figures/pipeline}
\input{figures/state_example}
\else

\input{figures/state_example}
\input{figures/pipeline}
\fi
\section{Introduction}
The rapid advancement of large language models (LLMs) has driven a paradigm shift from passive text generators to active API-calling agents that interact with stateful environments~\citep{appworld, wang2024officebench, coevolve}. While these agents can fulfill complex user requests by executing API calls and reasoning over environmental feedback, deploying them in real-world scenarios requires them to robustly generalize to new environments with unseen APIs. If trained on a narrow set of APIs, agents inevitably overfit to specific domain logic. Therefore, achieving true generalization hinges on broad supervision. This demands vast amounts of high-quality trajectory data covering numerous APIs from a diverse spectrum of stateful environments, capturing the full multi-step interaction process—including user requests, sequences of interleaved read and write API calls, execution results, and final responses. Compared to standard instruction-tuning data, collecting these trajectories at such an immense scale and breadth is substantially harder. Nevertheless, this expansive supervision is crucial for teaching agents robust API usage, accurate state tracking, and the ability to transfer their skills zero-shot to novel environments.

Given the difficulty of manual data collection, LLM-based synthetic data generation has emerged as a natural alternative. However, many existing works~\citep{toolllm, agenttuning, apigen, covert, lu2026firefly} primarily focus on read-only APIs, failing to equip agents with the skills needed to navigate stateful environments with both read and write APIs. While some recent approaches do target stateful scenarios, they typically assume access to fully implemented environments pre-populated with rich, realistic underlying states~\citep{coevolve, trajectory2task}. This tight coupling between environment construction and data synthesis creates a severe scalability bottleneck. While building such infrastructure for a few target applications is manageable, expanding coverage to a broad spectrum of APIs requires massive engineering effort and, more challengingly, entails the daunting burden of curating the diverse state configurations needed to populate these environments. Therefore, overcoming this scalability barrier necessitates a transformative shift in how we synthesize stateful trajectory data.

In this work, we ask a fundamental question: \emph{Can we scale the breadth of stateful agentic supervision by decoupling data synthesis from the environment bottleneck?} Our key conjecture is that modern LLMs, having been pre-trained on vast corpora of world knowledge, can function as dynamic digital world models. Given API specifications, input arguments, task instructions, and prior interaction history, an LLM can synthesize realistic API execution results that accurately reflect underlying state changes. Building on this insight, we introduce a scalable paradigm for generating broad agentic supervision. Rather than requiring fully-executable environments to generate our training data, we use an LLM to simulate state-consistent API responses on-the-fly directly from these specifications. This paradigm turns the formidable challenge of expanding environment coverage into the lightweight task of defining API specifications.

To instantiate this paradigm, we propose \textsc{SimSAT}  (\emph{\textbf{Sim}ulated \textbf{S}tateful \textbf{A}gentic \textbf{T}rajectories}), an LLM-based pipeline designed to synthesize broad agentic supervision starting solely from a list of domain names. The pipeline operates in four stages, as illustrated in \Cref{fig:pipeline}. First, during \textit{API synthesis}, a suite of applications is constructed for each domain, with each application defined as a collection of API specifications. Second, during \textit{task synthesis}, realistic and diverse tasks solvable by these APIs are created, and invalid requests are subsequently filtered out by an automated critic. Third, during the core \textit{trajectory synthesis} stage, a teacher agent is paired with an API simulator. As the teacher iteratively issues API calls to solve the task, the simulator dynamically tracks state changes and generates coherent environmental feedback on-the-fly using the user task and prior simulation history as context, resulting in complete, multi-step trajectories produced entirely in simulation (see \Cref{fig:state_example} for an example). Finally, during \textit{trajectory filtering}, an automated judge evaluates the resulting trajectories for correctness, completeness, and redundancy, discarding sub-optimal samples to ensure a high-fidelity training dataset. By decoupling training data generation from the environment bottleneck, our fully automated \textsc{SimSAT} pipeline makes it easy to scale stateful agentic supervision with minimal human effort.

To validate the effectiveness of \textsc{SimSAT}, we constructed a large synthetic dataset comprising 13.5K agentic trajectories, derived from 1,062 APIs across 65 stateful applications covering 12 broad domains. We then fine-tuned models of varying sizes (1.7B–14B) on this data and evaluated them zero-shot on AppWorld~\citep{appworld} and OfficeBench~\citep{wang2024officebench}—two challenging stateful benchmarks featuring unseen environments that test agents on complex, multi-step information retrieval and state-changing tasks. Fine-tuning on our broad \textsc{SimSAT} dataset yields significant performance gains of up to 50.0\% on AppWorld and 36.7\% on OfficeBench. Ultimately, these results demonstrate that \textsc{SimSAT} provides a highly scalable paradigm for training robust, generalizable, stateful API-calling agents.

Our contributions are summarized below:
\begin{itemize}[itemsep=1mm,parsep=1pt,topsep=1pt,leftmargin=*]
 \item \textbf{A Context-Aware API Simulator:} We introduce an LLM-based API simulator that acts as a digital world model. By leveraging task context and interaction history, it tracks underlying state changes and generates coherent feedback on-the-fly, enabling the synthesis of complex, multi-step trajectories without real API execution.
\item \textbf{The \textsc{SimSAT} Pipeline:} We propose \textsc{SimSAT}, a fully automated pipeline that decouples broad agentic supervision from the environment construction bottleneck. Starting solely from domain names, it seamlessly synthesizes diverse API specifications, tasks, and trajectories.
\item \textbf{Generalization to Unseen Environments:}
We demonstrate that agents trained purely on \textsc{SimSAT} data robustly generalize to unseen APIs. Across models ranging
from 1.7B to 14B parameters, our synthetic data drives substantial gains of up to 50.0\% on AppWorld and 36.7\% on OfficeBench.
\end{itemize}

%% file: figures/pipeline.tex
\begin{figure*}[ht!]
    \centering
    \resizebox{\textwidth}{!}{%
    \begin{tikzpicture}[
        node distance=0.4cm and 0.6cm,
        >={Stealth[length=3mm]},
        every node/.style={font=\normalsize},
        stagebox/.style={
            rounded corners=4pt,
            minimum height=1.4cm,
            minimum width=3.6cm,
            align=center,
            text=white,
            font=\normalsize\bfseries,
            drop shadow={shadow xshift=0.5mm, shadow yshift=-0.5mm, opacity=0.3},
        },
        subbox/.style={
            rounded corners=3pt,
            minimum height=0.85cm,
            minimum width=3.0cm,
            align=center,
            font=\small,
            draw=gray!50,
            fill=white,
        },
        outputbox/.style={
            rounded corners=4pt,
            minimum height=1.4cm,
            minimum width=3.2cm,
            align=center,
            text=white,
            font=\normalsize\bfseries,
            fill=gray!55!black,
            drop shadow={shadow xshift=0.5mm, shadow yshift=-0.5mm, opacity=0.3},
        },
        docfield/.style={
            font=\small\ttfamily,
            anchor=west,
            inner sep=1pt,
        },
        doclabel/.style={
            font=\small\bfseries,
            text=gray!50!black,
            anchor=west,
            inner sep=1pt,
        },
        arrowlabel/.style={
            font=\normalsize,
            text=gray!70!black,
        },
        innerarrow/.style={
            ->, thick, gray!60!black, shorten >=1pt, shorten <=1pt,
        },
    ]


    \node[stagebox, fill=stage0] (s0) {API Synthesis};
    \node[subbox, below=0.45cm of s0] (s0a) {LLM creates applications\\ from domain names};
    \node[subbox, below=0.45cm of s0a] (s0b) {LLM drafts API specs\\ for each application};
    \draw[innerarrow] (s0a.south) -- (s0b.north);

    \node[stagebox, fill=stage1, right=5.7cm of s0] (s1) {Task Synthesis};
    \node[subbox, below=0.45cm of s1] (s1a) {Task generator LLM\\ composes tasks};
    \node[subbox, below=0.45cm of s1a] (s1b) {Judge LLM filters\\ out invalid tasks};
    \draw[innerarrow] (s1a.south) -- (s1b.north);


    \coordinate (rowtwo) at ([yshift=-1.6cm]s0b.south -| s0.west);
    \node[stagebox, fill=stage2, anchor=north west] (s2) at (rowtwo) {Trajectory Synthesis};
    \node[subbox, below=0.45cm of s2] (s2a) {Agent LLM issues\\API calls};
    \node[subbox, below=0.45cm of s2a] (s2b) {Simulator LLM\\synthesizes responses};

    \draw[->, thick, stage2!80!black] (s2a.west) -- ++(-0.2,0) |- (s2b.west);
    \node[arrowlabel, align=center, anchor=east]
        at ([xshift=-0.42cm]$(s2a.west)!0.5!(s2b.west)$) {API\\calls};

    \draw[->, thick, stage2!80!black] (s2b.east) -- ++(0.2,0) |- (s2a.east);
    \node[arrowlabel, align=center, anchor=west]
        at ([xshift=0.42cm]$(s2a.east)!0.5!(s2b.east)$) {Responses};

    \node[stagebox, fill=stage3, right=3.3cm of s2] (s3) {Trajectory Filtering};
    \node[subbox, below=0.45cm of s3] (s3a) {Judge LLM evaluates\\ trajectory quality};
    \node[subbox, below=0.45cm of s3a] (s3b) {High-quality\\ trajectories selected};
    \draw[innerarrow] (s3a.south) -- (s3b.north);

    \begin{scope}[on background layer]
        \node[fit=(s0)(s0a)(s0b), rounded corners=6pt, fill=stage0bg, inner sep=6pt] (s0bg) {};
        \node[fit=(s1)(s1a)(s1b), rounded corners=6pt, fill=stage1bg, inner sep=6pt] (s1bg) {};
        \node[fit=(s2)(s2a)(s2b), rounded corners=6pt, fill=stage2bg, inner sep=6pt] (s2bg) {};
        \node[fit=(s3)(s3a)(s3b), rounded corners=6pt, fill=stage3bg, inner sep=6pt] (s3bg) {};
    \end{scope}

    \node[
        draw=gray!60, rounded corners=3pt, fill=white, inner sep=5pt,
        anchor=north east, align=left, font=\small,
        label={[font=\normalsize\bfseries, text=gray!40!black, yshift=-2pt]above:Domain names},
    ] (seed) at ([xshift=-1.05cm]s0bg.north west) {%
        \begin{tabular}{@{}l@{}}
            \textcolor{stage0!60!black}{Document Authoring \& Conv.}\\[1pt]
            \textcolor{stage0!60!black}{Files \& Data Operations}\\[1pt]
            \textcolor{stage0!60!black}{Social \& Communication}\\[1pt]
            \textcolor{stage0!60!black}{E-Commerce \& Marketplace}\\[1pt]
            \textcolor{stage0!60!black}{Productivity \& Proj.\ Mgmt}\\[1pt]
            \textcolor{stage0!60!black}{Entertainment \& Media}\\[1pt]
            \textcolor{stage0!60!black}{Travel \& Transportation}\\[1pt]
            \textcolor{stage0!60!black}{Food \& Health}\\[1pt]
            \textcolor{stage0!60!black}{Developer \& Business Tools}\\[1pt]
            \textcolor{stage0!60!black}{Finance \& Payments}\\[1pt]
            \textcolor{stage0!60!black}{Learning \& Hobbies}\\[1pt]
            \textcolor{stage0!60!black}{Local \& Community}
        \end{tabular}%
    };

    \node[outputbox, anchor=west] (output) at ([xshift=1.5cm]s3bg.east|-s3) {Trajectory dataset\\ for training};

    \node[
        draw=gray!60,
        rounded corners=3pt,
        fill=white,
        inner sep=6pt,
        anchor=center,
        align=left,
        label={[font=\normalsize\bfseries, text=gray!40!black, yshift=-2pt]above:API specifications},
    ] (spec) at ($(s0bg.east)!0.5!(s1bg.west)$) {%
        \begin{tikzpicture}[inner sep=0pt]
            \node[font=\small\itshape, text=gray!60!black, anchor=north west] (desc) at (0,0) {Send money to a user.};
            \node[font=\small\ttfamily\bfseries, text=stage0!60!black, anchor=south, yshift=2pt]
                (name) at ([xshift=1.45cm]desc.north west) {send\_money};
            \draw[gray!40, dashed] ([yshift=-5pt]desc.south west) -- ++(2.9,0);
            \node[doclabel, below=10pt of desc.south west, anchor=north west] (inlabel) {Input:};
            \node[docfield, below=1pt of inlabel.south west, anchor=north west, text=gray!30!black] (in1)
                {receiver\_id: \textcolor{typecolor}{int}};
            \node[docfield, below=0pt of in1.south west, anchor=north west, text=gray!30!black] (in2)
                {amount: \textcolor{typecolor}{float}};
            \node[docfield, below=0pt of in2.south west, anchor=north west, text=gray!30!black] (in3)
                {note: \textcolor{typecolor}{str} = \textcolor{stringcolor}{""}};
            \node[doclabel, below=3pt of in3.south west, anchor=north west] (outlabel) {Output:};
            \node[docfield, below=1pt of outlabel.south west, anchor=north west, text=gray!30!black] (out1) {\{};
            \node[docfield, below=0pt of out1.south west, anchor=north west, text=gray!30!black, xshift=10pt] (out2)
                {"txn\_id": \textcolor{typecolor}{int},};
            \node[docfield, below=0pt of out2.south west, anchor=north west, text=gray!30!black] (out3)
                {"status": \textcolor{typecolor}{str}};
            \node[docfield, below=0pt of out3.south west, anchor=north west, text=gray!30!black, xshift=-10pt] (out4) {\}};
        \end{tikzpicture}%
    };

    \draw[->, thick, gray!70!black] (seed.east|-s0) -- (s0bg.west|-s0);

    \draw[->, thick, gray!70!black] (s0bg.east|-s0) -- (spec.west|-s0);
    \draw[->, thick, gray!70!black] (spec.east|-s1) -- (s1bg.west|-s1);

    \coordinate (lane) at ($(s0b.south)!0.5!(s2.north)$);
    \coordinate (w1) at ([xshift=0.8cm]s1bg.east|-s1);
    \coordinate (w2) at (w1|-lane);
    \coordinate (w3) at ([xshift=-0.8cm]s2bg.west|-lane);
    \coordinate (w4) at ([xshift=-0.8cm]s2bg.west|-s2);
    \draw[->, thick, gray!70!black, rounded corners=4pt]
        (s1bg.east|-s1) -- (w1) -- (w2) -- (w3) -- (w4) -- (s2bg.west|-s2);
    \node[arrowlabel, anchor=east] at ([xshift=-0.15cm]$(w3)!0.5!(w4)$) {Tasks};

    \draw[->, thick, gray!70!black] (s2bg.east|-s2) -- (s3bg.west|-s3)
        node[arrowlabel, midway, above] {Trajectories};
    \draw[->, thick, gray!70!black] (s3bg.east|-s3) -- (output.west);

    \end{tikzpicture}
    }%
    \vspace{-0.05in}
    \caption{Overview of the \textsc{SimSAT} pipeline for synthesizing stateful agentic trajectories. (1) \textit{API Synthesis:} Starting from a list of twelve domain names, the pipeline synthesizes a broad set of API specifications. (2) \textit{Task Synthesis:} A diverse set of complex tasks is generated based on these APIs. (3) \textit{Trajectory Simulation:} A teacher agent iteratively solves the tasks while an LLM simulator—acting as a digital world model—dynamically tracks state and provides coherent API responses. (4) \textit{Quality Filtering:} An automated judge filters the resulting trajectories to yield high-quality training samples.
    }
    \label{fig:pipeline}
    \vspace{-0.15in}
\end{figure*}
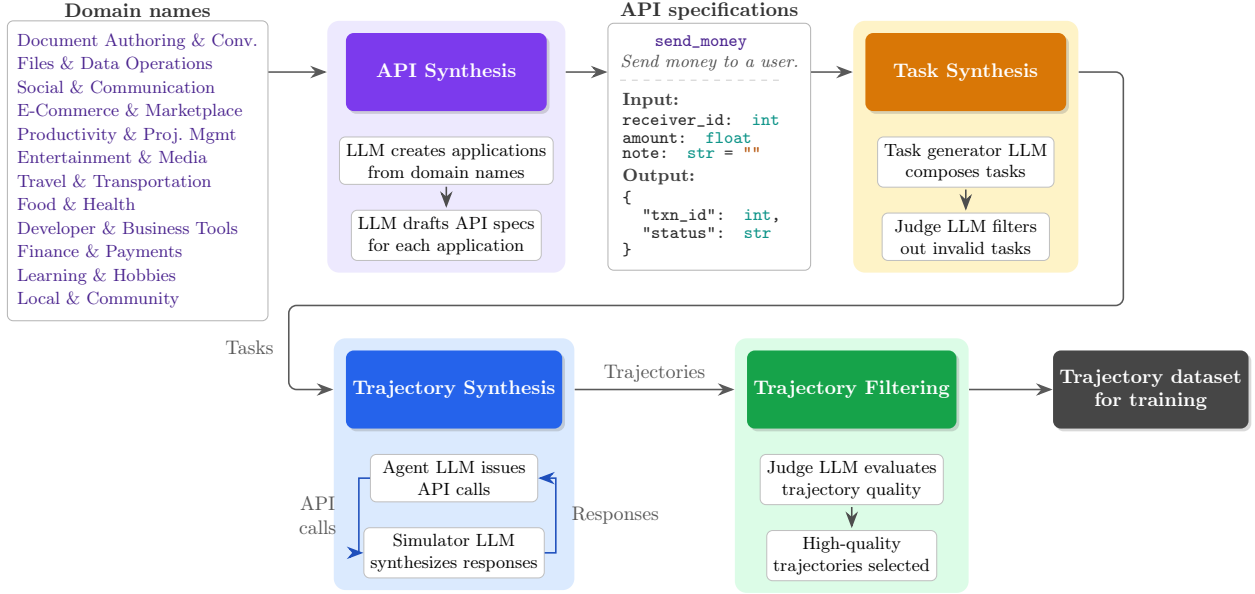

%% file: figures/state_example.tex
\providecommand{\trajcodesize}{\tiny}
\providecommand{\trajtextsize}{\scriptsize}
\begin{figure}[ht!]
\centering
\begin{trajboxf}{Task}
\trajtextsize Clean up my $\langle\mathtt{food\_delivery\_app}\rangle$ cart by
removing any item that costs more than \$20.\;\dots
\end{trajboxf}

\begin{agentboxf}{Agent}
\begin{Verbatim}[fontsize=\trajcodesize, breaklines, breakanywhere]
app = apis.<food_delivery_app>
tok = app.login(email=..., password=...)['access_token']
print(app.show_cart(access_token=tok))
\end{Verbatim}
\end{agentboxf}
\begin{envboxf}{Simulated API Response}
\begin{Verbatim}[fontsize=\trajcodesize, breaklines, breakanywhere]
{'items': [{'id': 401, 'name': 'Ribeye', 'price': 28.99, 'qty': 1},
           {'id': 405, 'name': 'Salmon', 'price': 24.50, 'qty': 1},
           {'id': 412, 'name': 'Burger', 'price': 14.99, 'qty': 2},
           {'id': 418, 'name': 'Salad',  'price': 11.99, 'qty': 1},
           {'id': 425, 'name': 'Pasta',  'price': 16.99, 'qty': 1}],
 'subtotal': 112.45, 'fee': 3.99, 'tax': 9.56, 'total': 126.0}
\end{Verbatim}
\end{envboxf}

\begin{agentboxf}{Agent}
\begin{Verbatim}[fontsize=\trajcodesize, breaklines, breakanywhere]
for i in [401, 405]:  # the two items priced over $20
    print(app.remove_from_cart(cart_item_id=i, access_token=tok))
\end{Verbatim}
\end{agentboxf}
\begin{envboxf}{Simulated API Response}
\begin{Verbatim}[fontsize=\trajcodesize, breaklines, breakanywhere]
{'message': 'Item removed.', 'cart_total': 94.54}
{'message': 'Item removed.', 'cart_total': 67.96}
\end{Verbatim}
\end{envboxf}

\begin{agentboxf}{Agent}
\begin{Verbatim}[fontsize=\trajcodesize, breaklines, breakanywhere]
print(app.show_cart(access_token=tok))
\end{Verbatim}
\end{agentboxf}
\begin{envboxf}{Simulated API Response}
\begin{Verbatim}[fontsize=\trajcodesize, breaklines, breakanywhere]
{'items': [{'id': 412, 'name': 'Burger', 'price': 14.99, 'qty': 2},
           {'id': 418, 'name': 'Salad',  'price': 11.99, 'qty': 1},
           {'id': 425, 'name': 'Pasta',  'price': 16.99, 'qty': 1}],
 'subtotal': 58.96, 'fee': 3.99, 'tax': 5.01, 'total': 67.96}
\end{Verbatim}
\end{envboxf}
\caption{\textbf{An example \textsc{SimSAT} trajectory:} Our API simulator generates a shopping cart on the fly when the \texttt{show\_cart} API is called, accurately simulates item deletions during subsequent \texttt{remove\_from\_cart} API calls, and reflects these state changes in the final shopping cart when 
\texttt{show\_cart} is invoked again.}
\label{fig:state_example}
\vspace{-16pt}
\end{figure}

%% file: sections/3_related_work.tex
\section{Related Work}

\paragraph{Stateless and read-only API agents.}
A large body of research on API-calling agents focuses on independent API calls or multi-step call sequences that do not require maintaining mutable shared state. Some approaches generate grounded supervision by executing calls against live APIs or executable functions~\citep{toolllm, agenttuning, covert, apigen, lu2026firefly}. Others synthesize API-calling trajectories using LLMs to simulate API responses~\citep{toolalpaca, toolace}. While these methods provide broad API coverage, they primarily address retrieval-oriented interactions rather than complex read-write workflows where underlying state changes must be dynamically tracked and reflected across subsequent calls.

\paragraph{Stateful agents in executable environments.}
A second line of work studies API-calling agents in fully implemented environments with persistent databases and state-changing APIs. Evaluation benchmarks such as AppWorld and OfficeBench require agents to perform long-horizon workflows involving interdependent read and write operations~\citep{appworld, wang2024officebench}. More recent training-oriented methods synthesize data within such environments. \citet{coevolve} generate tasks based on an agent's evolving weaknesses, validate them through environment interaction, and train the agent on the resulting tasks. \citet{trajectory2task} exploit executable APIs to obtain valid trajectories and convert them into verifiable user tasks. Although executable environments provide faithful state transitions and execution-based verification, their coverage remains bounded by the API suites, backend logic, and state configurations that have already been implemented. Extending these approaches to new API ecosystems therefore requires substantial engineering effort and the massive burden of curating realistic underlying states.

\paragraph{LLM-based simulation for stateful APIs.}
The most closely related works use LLMs to simulate stateful APIs. \citet{li2025simulating} augment seed trajectories for supervised fine-tuning and provide simulated environment feedback during reinforcement learning (RL). SynthAgent~\citep{synthagent} jointly synthesizes task-specific APIs, underspecified user requests, and simulated API responses for agentic RL. However, to stabilize repeated rollouts, SynthAgent relies on a task-level cache that maps API calls to static responses. Consequently, identical read calls receive the same cached response, preventing the simulator from accurately reflecting state changes caused by intervening write operations. Nemotron~\citep{chandiramani2026nemotron} generates large-scale simulated API-calling data through specialized and general-purpose pipelines. Its general-purpose pipeline constructs API suites from existing datasets and manually written APIs, leaving its coverage dependent on a pre-existing API pool. Furthermore, its generated trajectories average only five steps, limiting exposure to long-horizon interactions, and its from-scratch synthesis is restricted solely to customer-service scenarios. In contrast, \textsc{SimSAT} decouples data synthesis from existing infrastructure, synthesizing broad API suites and corresponding tasks directly from domain descriptions without relying on pre-existing API pools or seed trajectories. Furthermore, by conditioning each response on the preceding simulation history, our simulator acts as a digital world model that explicitly tracks state consistency across long-horizon trajectories with interleaved read and write calls.

%% file: sections/2_method.tex
\section{Method}
In this section, we present \textsc{SimSAT}, a fully automated pipeline that transforms a list of broad domain names into a massive dataset of multi-step, stateful API-calling trajectories. By decoupling data synthesis from environment construction, \textsc{SimSAT} bypasses the need for real API execution, backend infrastructure, or human-written tasks. As illustrated in \Cref{fig:pipeline}, the pipeline operates in four stages: (1) \textit{application synthesis}, which produces the API specifications that seed the pipeline; (2) \textit{task synthesis}, which formulates complex tasks based on those specifications; (3) \textit{trajectory synthesis}, where a teacher agent interacts with an API simulator acting as a digital world model; and (4) \textit{trajectory filtering} via an automated judge.

\subsection{Application Synthesis}
\paragraph{Application generation.}
We begin with a fixed taxonomy of twelve domain names (listed in \Cref{tab:domains}), ranging from finance and travel to productivity and document authoring. We provide the application synthesis LLM with a single example entry to demonstrate the required structure, which includes an application name, a short description, and a list of key agent-facing functionalities (see \Cref{lst:seed_domain}). Conditioned on this taxonomy and structural example, the model synthesizes diverse applications across the twelve domains. This process yields a total of 65 distinct applications.

\paragraph{API specification generation.}
For each application, we prompt the LLM to design a suite of API specifications with broad functional coverage. These include create-read-update-delete operations over the application's core resources, nested sub-resources, stateful actions, filtered list endpoints, and a set of account-management APIs.

\paragraph{Verification.}
Each generated specification is checked by a rule-based verifier to ensure that it contains well-defined input parameters and a valid response schema. Specifications that fail any check are discarded and regenerated. The resulting suite comprises 1,062 APIs across 65 applications, all of which are completely disjoint from the applications used in our downstream evaluation benchmarks. \Cref{sec:synthetic_apps} details the domain taxonomy, prompts, and verification rules.

\subsection{Task Synthesis}
This stage generates diverse user tasks solvable using the available APIs. The sole required input is a set of API specifications detailing each API's name, purpose, and input/output schemas.

\paragraph{Bucketized generation.} To ensure broad and balanced coverage, task generation is organized into a combinatorial grid of \emph{configuration buckets}. Each bucket is defined by a tuple of attributes, including \emph{difficulty level} (\eg, easy, medium, hard), \emph{action type} (\eg, read, write, mixed), \emph{task focus} (\eg, constraint satisfaction, derivation), the \emph{number of applications} (1, 2, or 3), and a \emph{range of APIs per application}. These buckets are processed in a randomized sequence, and a task generator LLM is prompted multiple times to yield a desired number of tasks per bucket. During each invocation, we randomly sample applications matching the bucket's required number and provide their full API specifications to the LLM, alongside the target difficulty, action type, task focus, and API count. The LLM then generates a task along with the list of APIs required to solve it. Detailed descriptions of the configuration attributes and their values are provided in \Cref{sec:bucket-gen}.

\paragraph{Inverse frequency sampling.} To ensure balanced coverage of applications and APIs, we employ inverse frequency sampling. We maintain running usage counts for every application and API across all generated tasks. When selecting applications for a generation step, we sample them with probabilities inversely proportional to their current counts, thereby prioritizing underrepresented applications. Furthermore, we rank all available APIs within the selected applications based on their counts, and explicitly instruct the task generator LLM to focus on the least-used APIs.

\paragraph{Task filtering.} Each generated task is evaluated by an LLM judge to ensure it is well-formed, solvable, and consistent with the corresponding bucket configuration.

\paragraph{Task rewriting.} Tasks generated directly from API specifications can sometimes suffer from procedural verbosity, outlining detailed instructions rather than natural goals. To address this, all tasks accepted by the judge are subsequently rewritten by an LLM to compress any procedural scaffolding into concise, intent-level requests that reflect actual user needs. These rewritten tasks are then verified by an LLM judge to ensure they remain valid. \Cref{sec:task_rewriting} provides examples illustrating the impact of this rewriting process. If a rewritten task is rejected by the judge, the corresponding initial task is retained for trajectory synthesis.

\subsection{Trajectory Synthesis}
In this stage, a teacher LLM agent iteratively solves each task by issuing API calls, while an LLM-based API simulator dynamically generates synthetic responses. This yields complete, multi-step training trajectories that include the agent's reasoning, API calls, and simulated environmental observations, all without executing any real APIs. 

To ensure structural correctness and semantic coherence, the simulator LLM conditions its output on a rich input context, which includes: (1) the full API specification, (2) the API call arguments, (3) the task being solved, (4) the history of previous API calls in the trajectory and their simulated responses, (5) the name, email, and phone number of a virtual user, and (6) the current date and time. To prevent overly long input contexts and accelerate simulation, we restrict the simulation history to previous API calls from the same application.

\paragraph{Rules for the simulator.} The simulator LLM is guided by six core rules encoded in its system prompt: (1) \emph{User-specific data}: responses must be consistent with the profile of the virtual user, who is treated as the logged-in user; (2) \emph{Schema compliance}: responses must strictly conform to the schema provided in the API specification, or to a fixed error object when the call is malformed or clearly infeasible given the preceding calls; (3) \emph{List generation}: the number of items returned must respect the pagination parameters and the entities established by earlier calls; (4) \emph{Data consistency}: it must preserve state across turns (\eg, a song assigned to a specific artist earlier must remain under that artist); (5) \emph{Data realism}: generated values must respect field constraints such as numeric ranges, timestamp ordering, and unique identifiers; and (6) \emph{Data diversity}: listed results should contain a mix of task-relevant and irrelevant items to mirror real-world noise.

\paragraph{Simulation quality checks.} To ensure that the simulator produces accurate and realistic responses, we employ multiple stages of quality control. Before invoking the simulator LLM, the API call arguments are programmatically validated for missing parameters and type mismatches. If errors are detected, the simulator directly returns an appropriate error message without invoking the LLM. For valid API calls, the LLM-generated response undergoes multiple checks. First, it is programmatically verified against the expected output schema. In the event of a schema mismatch, the simulator LLM is re-prompted using the schema errors as feedback. Once the response passes the schema check, it is evaluated for semantic correctness and consistency with the simulation history using the same LLM as a judge. If the judge identifies any errors, the simulator LLM is prompted to refine its response using the judge's output as feedback. Only responses that pass all checks are returned. We allow a fixed number of refinement steps. If the simulator fails to generate a valid response after exhausting all retries, we declare a simulation failure.

\paragraph{Trajectory termination.} Trajectory generation for a task concludes when the agent explicitly calls a special task completion API, exceeds a predefined limit on the number of execution steps, or invokes an API from an application that was not used to synthesize the corresponding task. The process also halts if the simulator fails.

\subsection{Trajectory Filtering}
Trajectories in which the agent completes the task by calling the task
completion API are evaluated by a trajectory-level LLM judge. The judge
determines whether the task was solved correctly. Each trajectory is
judged three times, and we retain only those with a majority-positive
verdict.

All the LLM prompts used for data synthesis are provided in~\Cref{sec:simulator-prompt}.

%% file: sections/4_experiment.tex
\section{Experiments}
\subsection{Experimental Setup}
\paragraph{Benchmarks.} We evaluate models trained on \textsc{SimSAT} data in a zero-shot fashion on two challenging, unseen benchmarks designed to test complex API interactions. \textbf{AppWorld} \citep{appworld} provides a testbed of intricate tasks that require agents to coordinate numerous read and write API calls across multiple applications. The environment features 457 APIs distributed across nine everyday applications, with the average task necessitating 10 unique APIs and 45 total calls. AppWorld provides two test splits: \textit{Test-Normal (Test-N)} with 168 tasks and \textit{Test-Challenge (Test-C)} with 417 tasks. \textbf{OfficeBench} \citep{wang2024officebench} focuses on office automation, featuring 20 APIs across eight applications. Tasks require agents to plan long-horizon workflows across applications, such as extracting spreadsheet data, scheduling calendar events, and dispatching emails. Following~\citet{li2025simulating}, we exclude OCR-dependent and unsolvable tasks, evaluating on 26 two-application and 46 three-application tasks.

\paragraph{Evaluation metrics.}
Following~\citet{appworld}, we report task and scenario goal completion (TGC and SGC) metrics on AppWorld. We report only TGC here and defer SGC to \Cref{sec:app-full-results}. For OfficeBench, we follow~\citet{wang2024officebench} and report the pass rate (\ie, the percentage of tasks passing their task-specific evaluation functions). All results reported in this section are averaged over eight inference runs, and the corresponding standard deviation values can be found in \Cref{sec:app-full-results}.

\paragraph{Models.} We used Gemini-2.5-Pro for API synthesis, GLM-4.7-FP8~\citep{glm-4.7} for task synthesis, GLM-5.1-FP8~\citep{glm} for trajectory synthesis (acting as both the teacher agent and API simulator), and Gemini-3.1-Pro~\citep{gemini31pro2026} for trajectory filtering. The resulting synthetic dataset was used to fine-tune Qwen3 (1.7B, 4B, 8B, and 14B) \citep{qwen3} and Qwen3.5 (2B, 4B, and 9B) \citep{qwen35} models.

\paragraph{\textsc{SimSAT} training data.} Using our pipeline, we synthesized 1,062 API specifications covering 65 applications across 12 broad domains, and generated trajectories with these APIs. Because AppWorld and OfficeBench require two different agent output formats (Python code in AppWorld and JSON-formatted tool calls in OfficeBench), we generated trajectories in both formats. Our \textsc{SimSAT} dataset comprises a total of 13,506 trajectories, with 6,265 following the Python code format and 7,241 following the JSON format. Detailed dataset statistics are provided in \Cref{app:stats-ESAT}.

\paragraph{Stateless simulator.} To isolate the contribution of context-aware state tracking, we additionally build a \textit{stateless} variant of our simulator that produces each API response from only the API specification and the call arguments, without access to the preceding call--response history. We run the identical synthesis pipeline with this simulator and fine-tune each model on the resulting trajectories using the same recipe, so the two conditions differ only in whether the simulator can condition on what has already happened in the trajectory.

\paragraph{Alternative synthetic datasets.} For comparison, we fine-tuned various Qwen models on five existing synthetic API-calling datasets: \textit{ToolAlpaca} \citep{toolalpaca}, \textit{ToolACE} \citep{toolace}, \textit{Nemotron}~\citep{chandiramani2026nemotron}, \textit{Simia}~\citep{li2025simulating}, and \textit{SynthAgent}~\citep{synthagent}. ToolAlpaca and ToolACE target simple API-calling scenarios and do not model stateful environments. While Nemotron contains stateful API-calling data, the scenarios are relatively simple and do not reflect the long-horizon, multi-application workflows present in AppWorld and OfficeBench. Simia simulates environment feedback to augment a small set of seed trajectories, and its released data is synthesized directly on top of the OfficeBench APIs; it is therefore in-domain for OfficeBench and covers no applications beyond those seeds. SynthAgent synthesizes tasks together with their tool ecosystems, but its mock tool system caches one static response per API call within a task, so repeated read calls return identical results and the resulting trajectories never exhibit the state changes induced by intervening write calls.

\paragraph{Training.} Each trajectory was formatted as a multi-step conversation consisting of the task instruction, the agent's reasoning and API calls, and the simulated API responses. We followed a standard supervised fine-tuning (SFT) strategy, training the model to predict the agent's output at each step conditioned on the task instruction and all previous API calls and their responses. Detailed training hyperparameters are provided in \Cref{sec:app-impl}.


\input{tables/appworld_main_results}
\subsection{Results on AppWorld}

\Cref{tab:appworld_main_results} reports the AppWorld performance of models fine-tuned on various synthetic datasets.

\paragraph{Large performance gains with \textsc{SimSAT}.} Fine-tuning on \textsc{SimSAT} data improves TGC by 13.9--49.3 points on Test-N and 6.2--50.0 points on Test-C over the corresponding zero-shot models. These gains are consistent across both model families and all parameter scales. In relative terms, the improvements are most dramatic for the smallest models, which are essentially unable to complete AppWorld tasks out-of-the-box. For instance, Qwen3-1.7B and Qwen3.5-2B jump from near-zero accuracy to 13.9 and 36.0 TGC on Test-N, respectively. The largest absolute gains appear at larger scales, with Qwen3.5-9B rising from 25.2 to 74.5 TGC on Test-N. Because the training data is generated entirely from synthetic applications, these results confirm that \textsc{SimSAT} data successfully imparts robust API-calling skills that transfer zero-shot to a completely unseen API ecosystem.

\paragraph{Limited effectiveness of existing synthetic datasets.} Fine-tuning on the five alternative synthetic datasets yields limited benefits and often degrades performance. Training on ToolAlpaca reduces performance below the zero-shot baseline for five of the seven models, while ToolACE does so for three. This degradation may reflect these datasets' emphasis on read-only APIs, which provides limited supervision for interacting with stateful environments. Simia and SynthAgent perform particularly poorly, achieving Test-N TGC scores of at most 6.5 and 14.7, respectively, across all seven models. Their poor performance may stem from different limitations: Simia's supervision is restricted to the OfficeBench applications used for synthesis, while SynthAgent's cached tool responses provide limited information about how state evolves over a trajectory. Nemotron, the strongest baseline, yields modest improvements for larger models, but \textsc{SimSAT} still outperforms it by substantial margins of 6.2--44.6 points. This gap suggests that Nemotron's relatively simple scenarios do not adequately prepare models for complex, long-horizon, multi-application workflows.

\paragraph{Small models surpassing much larger ones.} When fine-tuned on \textsc{SimSAT} data, several small models (Qwen3-4B, Qwen3-8B, Qwen3.5-4B, and Qwen3.5-9B) exceed the zero-shot performance of the significantly larger GPT-4o. Qwen3.5-4B is the most striking case: despite having only 4B parameters, it achieves 58.8 TGC on Test-C, nearly double GPT-4o's score of 30.2. This demonstrates the efficacy of our approach for knowledge distillation, confirming that complex, stateful API-calling capabilities can be successfully transferred to smaller models using purely simulated data.

\paragraph{Generalization to unseen user states.}
During trajectory generation, our API simulator dynamically simulates task-specific states for virtual users completely independent of AppWorld. The substantial performance gains observed on the AppWorld test splits indicate that models trained via \textsc{SimSAT} generalize not only to unseen APIs, but also to entirely unseen user states and underlying data distributions.

\input{tables/officebench}

\subsection{Results on OfficeBench}
\Cref{tab:officebench_main_results} presents the OfficeBench performance of models fine-tuned on \textsc{SimSAT} compared to alternative synthetic datasets.

\paragraph{Consistent gains on a second benchmark.}
Models fine-tuned on \textsc{SimSAT} data outperform the corresponding zero-shot baselines by 9.3--36.7 points on two-app tasks and 5.9--20.9 points on three-app tasks. These substantial gains once again confirm that our broad synthetic supervision imparts robust API-calling skills that successfully generalize to unseen environments.

\paragraph{Comparison with existing synthetic datasets.}
Fine-tuning on ToolAlpaca, ToolACE, and SynthAgent degrades performance for most models. Nemotron and Simia are the two stronger baselines: Nemotron yields modest gains for Qwen3.5-4B and Qwen3.5-9B, and Simia substantially helps the two smallest models on 2-app tasks. Simia's advantage here is expected, as its trajectories are synthesized directly on top of the OfficeBench APIs and are therefore in-domain, whereas \textsc{SimSAT} never observes an OfficeBench application during training. Even so, \textsc{SimSAT} outperforms Simia on thirteen of the fourteen model-split pairs, with Qwen3-1.7B 2-app as the only exception, and leads by 4.4--26.2 points on the harder 3-app split.

\paragraph{Effective distillation into compact models.}
Similar to the trends observed on AppWorld, fine-tuning on \textsc{SimSAT} data allows smaller models to exceed the performance of massive proprietary baselines. Specifically, the \textsc{SimSAT}-trained Qwen3.5-4B and Qwen3.5-9B models outperform zero-shot GPT-4o on OfficeBench. This reinforces the efficacy of our simulated data for distilling complex workflows into compact agents.

\subsection{Effect of Stateful Simulation}
\label{sec:stateless_ablation}
\Cref{tab:appworld_main_results,tab:officebench_main_results} also report models fine-tuned on trajectories from the stateless simulator. The full \textsc{SimSAT} simulator is better in all 28 model--split combinations, improving AppWorld TGC by 3.2--20.9 points on Test-N and 3.1--26.8 points on Test-C, and OfficeBench pass rate by 7.4--24.5 points on 2-app and 5.3--12.7 points on 3-app tasks. On OfficeBench the stateless variant is particularly unreliable, falling below zero-shot on 2-app tasks for Qwen3-8B and Qwen3-14B. Since the two conditions share an identical pipeline, this gap isolates the benefit of conditioning each simulated response on the preceding interaction history.

\subsection{Domain-Scaling Experiment}
\label{sec:domain_scaling}
\input{figures/domain_scaling}
A core premise of our work is that broad supervision leads to robust generalization. To empirically validate this, we examined how the breadth of synthetic training data affects downstream performance. We constructed nested training subsets by progressively increasing the number of domains in the order presented in \Cref{tab:domains}. To isolate breadth from data volume, every subset is subsampled to exactly 3,524 trajectories, so that all conditions share an identical training budget and differ only in how many domains that budget is spread across. We then fine-tuned the Qwen3.5-9B model on each subset. \Cref{fig:domain_scaling} illustrates how aggregate performance across the full AppWorld and OfficeBench test sets changes as domain coverage expands. Performance rises monotonically from 62.5\% when training on just three domains to 67.7\% when utilizing all twelve. This clear upward trend confirms our hypothesis: expanding the breadth of synthetic supervision enhances an agent's ability to generalize to unseen environments, and because the number of training trajectories is held fixed, the gain is attributable to breadth itself rather than to training on more data.



\subsection{Trajectory Filtering Quality}
Since our pipeline employs an LLM judge to filter synthetic trajectories, the quality of the final training dataset depends on the reliability of this judge. To validate its effectiveness, we conducted a human evaluation study. We randomly sampled 200 trajectories (100 accepted and 100 rejected by the judge) and instructed human annotators to assess them using the exact same grading criteria (see~\Cref{lst:judge_appworld_system}). The annotators disagreed with the judge on only 8 accepted and 2 rejected trajectories, yielding a 95\% overall agreement rate and a Cohen's kappa of 0.90. This near-perfect agreement confirms the high fidelity of our automated filtering process and, by extension, the quality of the resulting \textsc{SimSAT} dataset.

\subsection{API Simulator Quality Analysis}
\label{sec:simulator_quality}
While the previous sections establish the effectiveness of the \textsc{SimSAT} pipeline in terms of downstream model performance, this section evaluates the quality of the simulator as a standalone component. To this end, we sample 1,000 successful trajectories from our dataset, comprising 26,716 simulated API calls, and evaluate every simulated response using the latest proprietary model as a judge. For each API call, the judge was provided with the history of simulated call responses (which was used as the simulator's input context) and validated whether the current simulated response was consistent with the API specification, the call arguments, and the accumulated state from previous calls. Our simulator produced valid responses for 96.0\% of the calls, confirming the high quality of our \textsc{SimSAT} training dataset.

\paragraph{State tracking across write--read pairs.} The aggregate pass rate is dominated by calls whose responses can be produced without reasoning over the preceding trajectory. To measure state tracking directly, we isolate a subset of 469 read calls whose correct output depends on the effect of an earlier write call. The judge rates 88.7\% of these responses as both valid and consistent with the interaction history, indicating that the simulator propagates the state changes induced by write operations rather than returning values that merely look plausible for the API.

%% file: tables/appworld_main_results.tex
\begin{table*}[t!]
\centering
\footnotesize
\setlength{\tabcolsep}{2pt}
\begin{tabular}{l cc cc cc cc cc cc cc}
\toprule
& \multicolumn{2}{c}{\bf Qwen3-1.7B} & \multicolumn{2}{c}{\bf Qwen3-4B} & \multicolumn{2}{c}{\bf Qwen3-8B} & \multicolumn{2}{c}{\bf Qwen3-14B} & \multicolumn{2}{c}{\bf Qwen3.5-2B} & \multicolumn{2}{c}{\bf Qwen3.5-4B} & \multicolumn{2}{c}{\bf Qwen3.5-9B} \\
\cmidrule(lr){2-3} \cmidrule(lr){4-5} \cmidrule(lr){6-7} \cmidrule(lr){8-9} \cmidrule(lr){10-11} \cmidrule(lr){12-13} \cmidrule(lr){14-15}
& Test-N & Test-C & Test-N & Test-C & Test-N & Test-C & Test-N & Test-C & Test-N & Test-C & Test-N & Test-C & Test-N & Test-C \\
\midrule
zero-shot     & \phantom{0}0.0 & \phantom{0}0.1 & \phantom{0}8.7 & \phantom{0}3.5 & 20.2 & \phantom{0}7.8 & 33.2 & 16.3 & \phantom{0}0.1 & \phantom{0}0.2 & 27.8 & 17.3 & 25.2 & 15.4 \\
ToolAlpaca    & \phantom{0}0.1 & \phantom{0}0.2 & \phantom{0}6.3 & \phantom{0}2.0 & 16.9 & \phantom{0}4.9 & 25.5 & 12.8 & \phantom{0}0.2 & \phantom{0}0.3 & 25.7 & 17.3 & 24.4 & 15.7 \\
ToolACE       & \phantom{0}0.2 & \phantom{0}0.3 & \phantom{0}9.0 & \phantom{0}3.7 & 22.2 & \phantom{0}8.8 & 32.4 & 15.7 & \phantom{0}0.2 & \phantom{0}0.2 & 20.2 & 10.9 & 18.6 & 12.4 \\
Nemotron      & \phantom{0}0.0 & \phantom{0}0.1 & 12.4 & \phantom{0}7.1 & 26.8 & 14.4 & 42.7 & 30.0 & \phantom{0}0.5 & \phantom{0}0.2 & 26.0 & 17.5 & 31.9 & 20.8 \\
Simia         & \phantom{0}0.2 & \phantom{0}0.6 & \phantom{0}1.3 & \phantom{0}0.6 & \phantom{0}2.5 & \phantom{0}1.4 & \phantom{0}6.5 & \phantom{0}3.6 & \phantom{0}0.1 & \phantom{0}0.1 & \phantom{0}1.3 & \phantom{0}0.6 & \phantom{0}1.6 & \phantom{0}0.7 \\
SynthAgent & \phantom{0}0.2 & \phantom{0}0.2 & \phantom{0}2.5 & \phantom{0}1.0 & \phantom{0}2.8 & \phantom{0}1.3 & 11.5 & \phantom{0}6.8 & \phantom{0}0.0 & \phantom{0}0.1 & \phantom{0}9.2 & \phantom{0}4.4 & 14.7 & \phantom{0}6.3 \\
\midrule
Stateless     & \phantom{0}4.3 & \phantom{0}2.1 & 42.6 & 31.8 & 53.4 & 40.6 & 62.4 & 49.2 & 15.4 & \phantom{0}9.1 & 45.9 & 32.0 & 58.7 & 45.1 \\
\textsc{SimSAT} & \textbf{13.9} & \textbf{\phantom{0}6.3} & \textbf{50.3} & \textbf{37.2} & \textbf{56.6} & \textbf{43.7} & \bf 66.5 & \bf 53.7 & \bf 36.0 & \bf 27.0 & \bf 66.8 & \bf 58.8 & \bf 74.5 & \bf 65.4 \\
\midrule
\multicolumn{15}{@{}l@{}}{\begin{minipage}{\dimexpr\textwidth-2pt\relax}\footnotesize\vspace{2pt}
\emph{Off-the-shelf models (zero-shot) -- Test-N / Test-C:}~  GLM-5.1-FP8 -- 86.5 / 81.9;~ GPT-4o -- 48.8 / 30.2.\vspace{1pt}
\end{minipage}} \\
\bottomrule
\end{tabular}
\vspace{-0.1in}
\caption{TGC on AppWorld Test-N and Test-C for Qwen models fine-tuned on various synthetic datasets.}
\label{tab:appworld_main_results}
\end{table*}

%% file: tables/officebench.tex
\begin{table*}[t]
\centering
\footnotesize
\setlength{\tabcolsep}{2pt}
\begin{tabular*}{\textwidth}{@{\extracolsep{\fill}} l cc cc cc cc cc cc cc @{}}
\toprule
& \multicolumn{2}{c}{\bf Qwen3-1.7B} & \multicolumn{2}{c}{\bf Qwen3-4B} & \multicolumn{2}{c}{\bf Qwen3-8B} & \multicolumn{2}{c}{\bf Qwen3-14B} & \multicolumn{2}{c}{\bf Qwen3.5-2B} & \multicolumn{2}{c}{\bf Qwen3.5-4B} & \multicolumn{2}{c}{\bf Qwen3.5-9B} \\
\cmidrule(lr){2-3} \cmidrule(lr){4-5} \cmidrule(lr){6-7} \cmidrule(lr){8-9} \cmidrule(lr){10-11} \cmidrule(lr){12-13} \cmidrule(lr){14-15}
& 2-app & 3-app & 2-app & 3-app & 2-app & 3-app & 2-app & 3-app & 2-app & 3-app & 2-app & 3-app & 2-app & 3-app \\
\midrule
zero-shot     & 22.8 & \phantom{0}4.6 & 49.8 & 38.9 & 63.2 & 32.7 & 69.1 & 47.5 & \phantom{0}9.1 & \phantom{0}0.7 & 62.3 & 37.1 & 68.6 & 46.4 \\
ToolAlpaca    & \phantom{0}6.4 & \phantom{0}0.0 & 11.3 & \phantom{0}2.1 & 29.7 & 10.9 & 30.6 & 12.1 & \phantom{0}8.8 & \phantom{0}0.5 & 25.3 & \phantom{0}8.0 & \phantom{0}7.8 & \phantom{0}0.2 \\
ToolACE       & \phantom{0}5.9 & \phantom{0}0.0 & 15.4 & \phantom{0}3.0 & 29.2 & \phantom{0}5.7 & 37.3 & 22.5 & \phantom{0}5.9 & \phantom{0}0.0 & 10.3 & \phantom{0}5.2 & 36.3 & 11.4 \\
Nemotron      & \phantom{0}5.9 & \phantom{0}0.0 & \phantom{0}8.1 & \phantom{0}0.5 & 52.0 & 35.5 & 70.1 & 45.0 & 11.8 & \phantom{0}0.2 & 68.2 & 40.7 & 73.8 & 49.6 \\
Simia & \bf 40.4 & \phantom{0}6.1 & 55.2 & 25.2 & 61.3 & 31.1 & 65.4 & 33.9 & 43.4 & 16.4 & 64.7 & 29.6 & 71.6 & 38.0 \\
SynthAgent & \phantom{0}6.1 & \phantom{0}0.0 & 17.6 & \phantom{0}5.5 & 39.2 & 18.6 & 54.2 & 27.3 & 10.5 & \phantom{0}0.9 & 49.0 & 24.1 & 70.3 & 50.0 \\
\midrule
Stateless & 25.7 & \phantom{0}5.2 & 56.2 & 43.8 & 58.4 & 42.1 & 60.7 & 45.3 & 21.3 & \phantom{0}8.9 & 66.3 & 39.9 & 71.8 & 50.4 \\
\textsc{SIMSAT} & 36.3 & \bf 10.5 & \textbf{68.9} & \textbf{51.4} & \bf 78.2 & \bf 48.4 & \bf 78.4 & \bf 53.0 & \bf 45.8 & \bf 21.6 & \bf 79.2 & \bf 52.1 & \bf 79.2 & \bf 60.0 \\
\bottomrule
\end{tabular*}

\vspace{-1pt}
\begin{minipage}{\textwidth}\vspace{4pt}
\emph{Off-the-shelf models (zero-shot) -- 2-app / 3-app:}~GLM-5.1-FP8 -- 90.2 / 74.3;~ GPT-4o -- 74.5 / 50.9.\vspace{3pt}
\hrule height\heavyrulewidth
\end{minipage}
\vspace{-0.1in}
\caption{Pass rate  on OfficeBench 2-app and 3-app tasks for Qwen models fine-tuned on various synthetic datasets.}
\label{tab:officebench_main_results}
\end{table*}

%% file: figures/domain_scaling.tex
\begin{domainfig}
    \centering
    \includegraphics[width=\domainfigwidth]{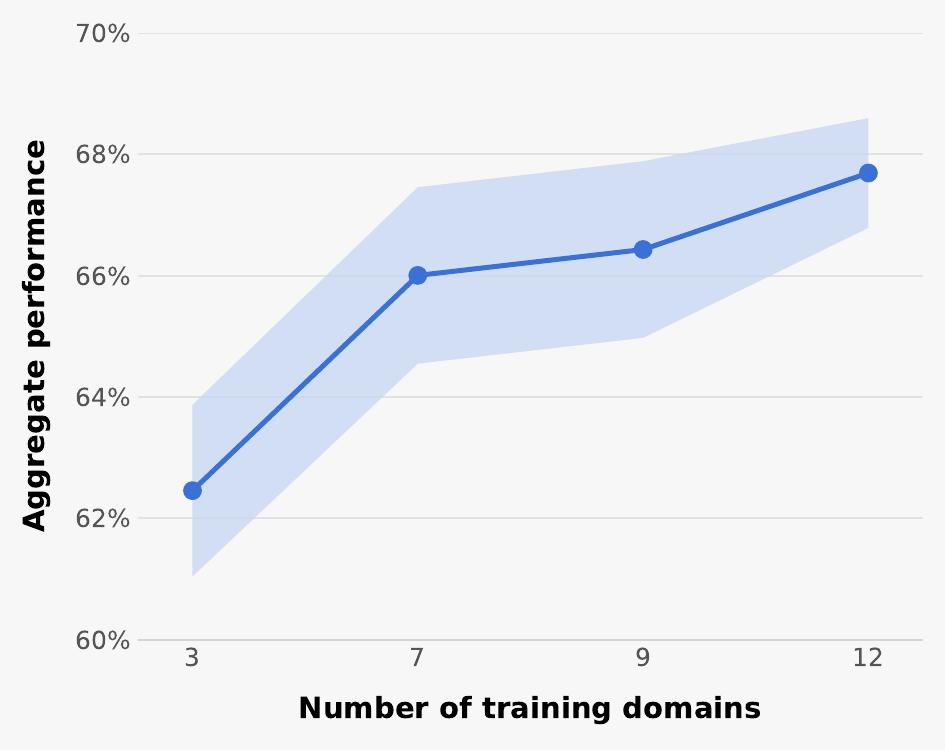}
    \vspace{-0.1in}
    \caption{Effect of scaling supervision breadth on the performance of Qwen3.5-9B model aggregated on the full AppWorld and OfficeBench test sets.}
    \label{fig:domain_scaling}
    \vspace{-0.05in}
\end{domainfig}

%% file: sections/5_conclusion.tex
\section{Conclusion}
We presented \textsc{SimSAT}, a fully automated pipeline that synthesizes training data for stateful API-calling agents starting solely from a list of domain names. By decoupling data synthesis from the environment construction bottleneck, \textsc{SimSAT} combines automated API synthesis, coverage-aware task generation, a state-consistent API simulator, and a trajectory-level judge to produce high-quality, multi-step trajectories. Evaluations on the challenging AppWorld and OfficeBench benchmarks demonstrate substantial zero-shot gains across models ranging from 1.7B to 14B parameters, even though the synthetic APIs used for training are completely disjoint from those in either benchmark. Compact models also outperform the massive GPT-4o baseline on both benchmarks: four models between 4B and 9B parameters exceed it on AppWorld, and Qwen3.5-4B and Qwen3.5-9B do so on OfficeBench. These results demonstrate that purely synthetic domains suffice to teach robust API-calling skills that transfer to unseen ecosystems, establishing our LLM world-model-based synthesis approach as a highly scalable solution for training generalizable agents.

%% file: sections/6_appendix.tex
\clearpage
\input{tables/task_throughput}

\section{Data Statistics of \textsc{SimSAT}}
\label{app:stats-ESAT}

\subsection{Synthesis Throughput and Yield}

Because AppWorld and OfficeBench expect different agent output formats, we ran the synthesis pipeline twice, once for the Python coding format and once for the JSON tool-calling format, and report the throughput of each run separately below. Note that when a rewritten task is rejected, its corresponding initial task is retained, so all tasks accepted by the first judge remain available for trajectory synthesis. Together the two runs supply the 13,506 trajectories of the \textsc{SimSAT} dataset.

\subsubsection{Python Coding Format}

\paragraph{Task synthesis throughput.}
\Cref{tab:syn_task_synthesis_python} outlines task retention at each phase of the pipeline. The first judge preserves 10,509 of 32,009 generated tasks (32.83\%), discarding roughly 70\% for being unsolvable, improperly formatted, or mismatched with their designated bucket attributes. The surviving tasks undergo a rewriting phase that turns each one into a concise, intent-focused request, which yields 9,171 successful reformulations (28.65\% of the initial pool). Finally, a secondary judge cross-checks the rewritten prompts against the original criteria, leaving a curated pool of 7,072 tasks (22.09\% of the initial count). Including the initial tasks retained for rejected rewrites, this run supplies a pool of 10,509 tasks for trajectory synthesis.

\input{tables/syn_throughput}
\paragraph{Trajectory generation throughput.}
As reported in \Cref{tab:syn_traj_througput_python}, the teacher agent completes 8,020 of the 10,509 input tasks (76.32\%), meaning the agent called a special task completion API irrespective of its actual correctness, and the judge retains 6,265 of these trajectories (78.12\% of the completed ones, or 59.62\% of the input tasks). Simulator retries are rare: only 6,062 of 154,317 simulated calls (3.93\%) required a retry due to schema or semantic validation errors, indicating that the simulator reliably generates well-formed and consistent responses on the first attempt.

\subsubsection{JSON Tool-Calling Format}

\paragraph{Task synthesis throughput.}
The JSON-format run follows the same pattern at a larger scale, as reported in \Cref{tab:syn_task_synthesis_stats_json}. From an initial batch of 50,343 generated tasks, 15,220 pass the initial judge (30.23\%), 13,876 are successfully rewritten (27.56\%), and 10,484 pass the final judge (20.83\%). This run supplies a pool of 15,220 tasks for trajectory synthesis.

\paragraph{Trajectory generation throughput.}
\Cref{tab:syn_traj_througput_json} shows that the agent completes 10,662 of the 15,220 input tasks (70.05\%) and that 7,241 trajectories survive filtering (47.58\% of the input tasks), a lower retention rate than in the Python format. The gap comes mostly from the filtering stage rather than from completion: the judge rejects 3,421 of the 10,662 completed trajectories (32.1\%), compared with 1,755 of 8,020 (21.9\%) in the Python format. We analyze the cause of these rejections in \Cref{app:failure-json}. Simulator retries remain rare, with 2,454 of 68,436 simulated calls (3.59\%) requiring a retry, a rate comparable to the Python-format run.

\subsection{Failure Analysis}
\input{tables/syn_failure_distribution}

\subsubsection{Python Coding Format}

\paragraph{Failure types in trajectory generation.}
\Cref{tab:syn_failure_distribution} breaks down the 4,244 discarded Python-format trajectories by failure mode. The dominant cause is \emph{judge rejection} (41.35\%). These trajectories are completed end to end but are filtered out by the LLM judge because their solutions are incorrect or of low quality. Thus, this category reflects the pipeline's intended quality-control mechanism rather than a pipeline error. The second-largest category is \emph{simulation failure} (28.30\%), which occurs when the simulator cannot produce a response that passes both the schema check and the LLM-as-a-judge consistency check within the retry budget, or when it times out because the agent issues too many simulated API calls in a single step. \emph{Agent's unexpected App use} (22.05\%) occurs when the agent calls an API outside the expected set derived from the task, prompting early termination. \emph{Agent model execution failures} (8.25\%) are caused by transient errors in the teacher LLM serving stack, such as rate-limit or decoding errors. \emph{Max solve steps exceeded}, which occurs when the agent fails to terminate within the step budget, is negligible for this format (0.02\%). A single \emph{unexpected} runtime error (0.02\%) accounts for the remainder.

\subsubsection{JSON Tool-Calling Format}
\label{app:failure-json}

\paragraph{Failure types in trajectory generation.}
\Cref{tab:syn_failure_distribution_json} breaks down the 7,979 discarded JSON-format trajectories by failure mode. As in the Python format, \emph{judge rejection} is the dominant category, and at a comparable rate (42.88\%, versus 41.35\%). The two formats diverge sharply elsewhere: \emph{max solve steps exceeded} accounts for 25.76\% of JSON-format failures but only 0.02\% of Python-format ones. We attribute this to the relative expressiveness of the two action formats. A single Python step can execute an entire complex procedure using loops, conditional branching, and intermediate computations over the results of earlier calls. By contrast, the JSON tool-calling format restricts each step to a single API call, with no intermediate computation. Many of our tasks require precisely this type of composite processing. In the JSON format, the agent must instead unroll the procedure across many turns, tracking and combining raw API responses in its context rather than processing them programmatically, and for a substantial fraction of tasks that unrolled sequence does not fit within the step budget. Where it does fit, the agent may still drop constraints, incorrectly aggregate intermediate values, rely on stale values, or take shortcuts during long iterations. Such trajectories are complete but incorrect and appear as judge rejections rather than timeouts. Conversely, \emph{simulation failure} is a much smaller category in this format (9.06\%, compared with 28.30\% in the Python format). \emph{Agent's unexpected App use} (20.68\%) is broadly comparable to that observed in the Python-format run, while \emph{agent model execution failures} are less frequent (1.63\%). No \emph{unexpected} runtime errors occurred.

\subsection{Dataset Diversity and Coverage}
\input{figures/syn_task_dist}

\paragraph{Task distribution.}
\Cref{fig:syn_task_focus,fig:syn_action_type,fig:syn_difficulty,fig:syn_num_apps} present the distribution of the generated tasks used for trajectory synthesis across four dimensions. The generated tasks exhibit fairly uniform coverage across task focus categories (24--27\%) and action types (30--36\%), and are moderately tilted toward harder tasks (easy 29\%, medium 34\%, hard 37\%). The distribution is most skewed with respect to the number of apps involved: single-app tasks make up the majority at 57\%, with two-app and three-app tasks comprising 29\% and 14\% of the pool, respectively. This could be due to the inherent difficulty an LLM-based task generator faces when composing meaningful multi-app tasks. Additionally, it is likely that this distribution reflects a naturally non-uniform distribution of tasks in the real world.

\paragraph{Trajectory distribution.}
To ensure our synthetic dataset captures a diverse and balanced range of agentic capabilities, we analyze the distribution of the successfully generated trajectories across four dimensions, as illustrated in \Cref{fig:syn_traj_task_focus,fig:syn_traj_action_type,fig:syn_traj_difficulty,fig:syn_traj_num_apps}. The dataset exhibits broad coverage across task difficulty levels (easy 31\%, medium 32\%, hard 38\%), action types (mixed 41\%, write 30\%, read 29\%), and task focus categories (constraint satisfaction 23\%, derivation 24\%, iteration 25\%, open 28\%). As in the task pool, however, the distribution is skewed regarding the number of apps involved, with two-app (20\%) and three-app (6\%) trajectories being substantially less frequent than single-app trajectories (74\%). This skew is more pronounced than in the task pool, indicating that multi-app tasks are also harder for the teacher agent to complete successfully.

\paragraph{API coverage.}
To assess the diversity and breadth of our synthesized dataset, we measure API coverage by tracking the number of unique APIs successfully invoked throughout the generated trajectories. We determine this coverage using various frequency thresholds, calculating the proportion of APIs that appear at least $N$ times (where $N \in \{1, 2, 3, 5, 10, 20, 50\}$). As illustrated in \Cref{fig:syn_api_coverage}, our pipeline achieves near-complete coverage across the environment. Out of the APIs distributed across our 65 synthetic apps, 1,058 (99.6\%) are called in at least one successful trajectory, and 996 (93.8\%) are utilized at least five times. These high coverage rates confirm that our inverse-frequency sampling strategy effectively prevents the data distribution from collapsing onto a narrow subset of popular endpoints, ensuring the agent develops robust and generalized API-calling capabilities.

\input{figures/syn_api_coverage}

\section{Implementation Details}
\label{sec:app-impl}

\paragraph{Data synthesis.}
For API synthesis (Stage 1), we use Gemini-2.5-Pro to expand the seed domains and to generate the API specification for each application. For task synthesis (Stage 2), we use GLM-4.7-FP8 as both the task generator and the LLM judge that validates generated tasks. For trajectory synthesis (Stage 3), GLM-5.1-FP8 serves as the teacher agent and the API simulator, and the maximum number of agent steps per trajectory is set to 50. Trajectories that exceed this limit are aborted as described in the trajectory termination rules. For trajectory filtering (Stage 4), we use Gemini-3.1-Pro as the LLM judge.

\paragraph{Data volume and synthesis cost.}
The final \textsc{SimSAT} training corpus contains 148,702,719 tokens. In comparison, the Nemotron dataset used in our controlled baseline contains 1.67B tokens, approximately 11.2 times more. Thus, the performance advantage of \textsc{SimSAT} cannot be attributed to a larger training-token budget. The pipeline is also inexpensive to run, as it requires only API calls and no executable environment to build or maintain. At the prices of the models we used, application synthesis costs \$0.09 per application, task synthesis \$0.04 per task, and trajectory generation \$0.30 per trajectory. Filtering costs \$0.08 per judge call, and we query the judge three times per trajectory, adding \$0.24. Synthesizing the full suite of 65 applications therefore costs under \$6, and the marginal cost of generating and filtering a single trajectory is \$0.54.

\paragraph{Training.}
We fine-tune all models with supervised learning on full multi-turn trajectories. Each trajectory is decomposed into individual turns, and the model is trained to predict the tokens of the current turn conditioned on all preceding turns, with the loss masked out over the context tokens. We use the AdamW~\citep{adamw} optimizer with a learning rate of $2\cdot 10^{-5}$, a linear decay schedule, and a warmup ratio of 1\%. Training is performed with a context length of 32{,}768 tokens and a per-device batch size of 1. For models smaller than 14B parameters, training is distributed across 8 B200 GPUs. We leverage the Transformers library~\citep{wolf-etal-2020-transformers} for model implementation and use PyTorch native FSDP2 for distributed training. To improve training efficiency, we apply sequence packing, which concatenates roughly 4 samples into a single sequence on average. We train for 10 epochs by default, except for the Qwen3.5 family, where we observed overfitting beyond 5 epochs and therefore reduce training to 5 epochs.

\paragraph{Inference and evaluation.}
For efficient, high-throughput inference, we evaluate all models using vLLM~\citep{kwon2023efficient}. During evaluation, we set the sampling temperature to 1.0. Following the evaluation protocol of AppWorld and OfficeBench, the maximum number of agent steps is capped at 50 for AppWorld and 30 for OfficeBench. For both benchmarks, we generate 8 independent samples per task and report the average pass@1 metric to robustly measure agent performance.

\section{Full Experimental Results}
\label{sec:app-full-results}
We present the complete evaluation results for our experiments in this section. \Cref{tab:appworld-app} details the comprehensive performance on the AppWorld benchmark, reporting both Task Goal Completion (TGC) and Scenario Goal Completion (SGC) metrics across the Test-Normal and Test-Challenge splits. Similarly, \Cref{tab:officebench_std} provides the full pass rate breakdown for 2-app, 3-app, and overall tasks on the OfficeBench dataset. For both benchmarks, we report the mean performance along with standard deviations across 8 independent inference runs for all evaluated models and training data configurations.

\input{tables/appworld_appendix}

\input{tables/officebench_app}

\subsection{Effect of the Teacher Agent}
\label{app:teacher_ablation}
Our main pipeline uses GLM-5.1-FP8 as both the teacher agent and the API simulator. To test whether the pipeline depends on access to a frontier teacher, we regenerate the dataset with the weaker GLM-4.7 acting as the teacher agent while keeping GLM-5.1-FP8 as the simulator, and fine-tune each model on the result using the identical recipe. \Cref{tab:teacher_appworld,tab:teacher_officebench} report AppWorld and OfficeBench performance for both teachers alongside the zero-shot baseline. Even with the weaker teacher, every model improves substantially over its zero-shot baseline on both benchmarks, so the effectiveness of the pipeline is robust to the choice of teacher rather than an artifact of distilling one particular frontier model. Upgrading the teacher to GLM-5.1-FP8 improves results further in 24 of the 28 cells, showing that the approach also scales as stronger teachers become available. The exceptions are the two smallest models, Qwen3-1.7B on AppWorld and Qwen3.5-2B on OfficeBench, for which student capacity rather than demonstration quality is the binding constraint.

\input{tables/teacher_appworld}

\input{tables/teacher_officebench}

\subsection{Effect of Explicit Reasoning in the Training Data}
\label{app:explicit_reasoning}
Trajectories synthesized by our pipeline interleave the teacher agent's explicit reasoning with its API calls. To measure how much that reasoning contributes, we remove the reasoning from every agent turn, leaving an empty \texttt{<think></think>} block so that the structure of the turn is preserved, and fine-tune Qwen3.5-9B on the result with the same recipe. \Cref{tab:explicit_reasoning} reports the outcome. Removing the reasoning costs 18.4 and 20.5 TGC on AppWorld Test-N and Test-C, and 14.0 and 17.7 points on OfficeBench 2-app and 3-app tasks. The supervision therefore derives much of its value from the demonstrated reasoning rather than from the action sequences alone, and the loss is larger on the harder split of each benchmark, where choosing the next call depends on more of the preceding context.

\input{tables/explicit_reasoning}

\section{Task Synthesis}

\label{sec:task_synthesis}
The goal of task synthesis is to produce a large and diverse pool of natural, multi-step user tasks that can be solved using only the available APIs. The only input to this stage is a set of API specifications, where each specification provides an API's name,
purpose, input-parameter schema, and output-response schema. From these specifications, we generate, judge, and rewrite candidate tasks. We use a single LLM, GLM-4.7-FP8~\citep{glm-4.7}, for all three roles (generator, judge, and rewriter) by varying the prompt.

\subsection{Bucketized Generation}
\label{sec:bucket-gen}
To ensure broad and balanced coverage of the task space, generation is organized into a combinatorial grid of \emph{configuration buckets}. Each bucket is a tuple $(\text{difficulty},\allowbreak\ \text{action\_type},\allowbreak\ \text{task\_focus},\allowbreak\ \text{num\_apps},\allowbreak\ [\text{min\_apis},\text{max\_apis}])$, where (i) \emph{difficulty} $\in\{\textsc{easy},\textsc{medium},\textsc{hard}\}$ controls the depth of reasoning required (\Cref{tab:taskgen_difficulty}); (ii) \emph{action\_type} $\in\{\textsc{read},\textsc{write},\textsc{mixed}\}$ controls the mix of \texttt{GET} vs.\ \texttt{POST}/\texttt{PATCH}/\texttt{DELETE} calls (\Cref{tab:taskgen_action}); (iii) \emph{task\_focus} (\eg, constraint satisfaction, derivation, iteration, open) controls the high-level reasoning pattern (\Cref{tab:taskgen_focus}); (iv) \emph{num\_apps} $\in\{1,2,3\}$ specifies the number of distinct apps a task must span; and (v) $[\text{min\_apis},\text{max\_apis}]$ specifies the per-app API budget. The per-app API budgets are pinned to match the typical per-app load needed at each app count: for $\text{num\_apps}=1$, we sweep five pinned values $[k,k]$ for $k\in\{2,\dots,6\}$; for $\text{num\_apps}=2$, we use ranges $\{[1,2],[2,3],[3,4]\}$; and for $\text{num\_apps}=3$, we use $\{[1,2],[2,3]\}$. This yields $3\cdot3\cdot4\cdot(5+3+2)=360$ buckets in total. Buckets are processed sequentially in a random order, and we generate $N_{\text{bucket}}=100$ candidate tasks for each bucket.

For each bucket, the pipeline issues 10 parallel calls to the task generator LLM, each asked to produce $n_{\text{per\_call}}=10$ distinct candidate tasks. The resulting candidates are subsequently evaluated by the task judge. For hard buckets ($\text{difficulty}=\textsc{hard}$, $\text{num\_apps}=3$, or $\text{num\_apps}\cdot\text{min\_apis}>5$), we double the maximum completion-token budget for the generator, judge, and rewriter to accommodate longer outputs.

\subsection{Inverse Frequency Sampling}
To prevent the generator from collapsing onto a small set of apps and APIs, we maintain running usage counts $c(a)$ and $c(\alpha)$ for every app $a$ and every API $\alpha$ across all tasks generated so far. Coverage is refreshed at the start of each bucket from this
job's own outputs. When constructing a generator call:
\begin{enumerate}[leftmargin=*,topsep=2pt,itemsep=2pt]
\item \textbf{App sampling.} We sample $\text{num\_apps}$ apps without replacement using
$p(a)\propto 1/(1+c(a)),$
which assigns the highest probability to the least-used apps.

\item \textbf{API Spotlighting.} For each selected app, we provide the generator with the full set of API specifications for that app, but additionally surface a list of the top-10 APIs ranked by inverse coverage as \emph{focus APIs}. These are listed under a dedicated section of the prompt with instructions to prioritize them, while still leaving the generator free to use any of the provided APIs to satisfy other requirements.

\item \textbf{Prompt assembly.} The selected app/API specification, the focus-API list, the bucket-specific difficulty, action-type, task-focus clauses (\Cref{tab:taskgen_difficulty,tab:taskgen_action,tab:taskgen_focus}), and the per-app API-count range are inserted into the task generation prompt template shown in \Cref{lst:taskgen_prompt}.
\end{enumerate}

\subsection{Task Judging}
Each surviving candidate is then evaluated by an LLM judge that is given the same API specification and the exact requirements section that was shown to the generator (the universal requirements plus the bucket-specific difficulty, action-type, task-focus, coverage, and required-apps clauses). \Cref{lst:judge_prompt} shows the full prompt. The judge is instructed to think through how the task would be solved, since this is necessary for it to assess API coverage, difficulty, and focus compliance, and then output a single yes or no verdict. Tasks receiving a No are excluded from downstream stages.

\subsection{Task Rewriting}
\label{sec:task_rewriting}
Tasks generated directly from API specifications often suffer from \emph{procedural verbosity}: the generator tends to enumerate intermediate steps that any agent would infer on its own (\eg, ``first look up X, then check Y, then if Z\dots''). To convert these into
natural, intent-level user requests, every task that passes the judge is rewritten by a second LLM call. The rewriter is shown (i) the original task, (ii) the same task-requirements block used by the generator and judge, and (iii) only the API specification of the APIs
covered by the task (rather than the full bucket-level pool) so that it stays focused on preserving the original API coverage. To anchor the desired writing style, we append three in-context examples to the prompt body that pair a verbose, procedural original with its
compact, goal-oriented rewrite (\Cref{lst:rewrite_prompt}, \Cref{lst:rewrite_examples}). The rewriter is explicitly instructed to preserve every detail the user actually cares about, while stripping setup phrases and obvious intermediate steps. \Cref{tab:rewrite_examples} shows two representative before-and-after pairs. In both cases, the initial generated task reads as a detailed, multi-sentence walkthrough that explicitly enumerates the steps the agent should take. The rewritten task preserves the same underlying intent, but expresses it as a short, natural user request that omits the procedural scaffolding.

\subsection{Final Re-judging}
Because rewriting can occasionally omit a required entity or alter the task's scope, every rewritten task is evaluated by the judge a second time using the same prompt template. If the rewritten task passes this second evaluation, it is used for trajectory synthesis. Otherwise, we fall back to the corresponding original task, which has already passed the initial judging stage. Thus, every task accepted by the initial judge proceeds to trajectory synthesis, either in its rewritten form or, when rewriting is unsuccessful, in its original form.

\input{tables/task_examples}


\input{tables/task_prompt}

\section{Simulator Details}
\label{app:simulator}
This section provides a detailed description of the LLM-based API simulator used during trajectory synthesis. The simulator is purely text-based: there is no real backend, no database, and no network call. Its job is to take an API call issued by the teacher agent and
produce a JSON response that (i) conforms to the documented response schema, (ii) is internally consistent with all responses returned so far in the same task, and (iii) plausibly reflects the user profile and the current task.

\subsection{Per-Task, Per-App Histories}
\label{app:sim_history}
For each task, the simulator maintains an independent rolling history of accepted (call, response) pairs for each app. Restricting the history to a single app sharply reduces prompt length and prevents irrelevant entities from leaking into one app's simulated state.

When constructing a prompt, the simulator formats the most recent calls in this app's history as ``Call $1$ \dots Call $N$,'' each entry containing the API signature, the input parameters, and the previously simulated response. The history is capped at a
configurable maximum number of calls, but in practice the per-task length stays well below this cap. When a task finishes, its histories are explicitly cleared, so simulator state never leaks between tasks.

\subsection{Deterministic Input Validation}
\label{app:sim_validation}
Inputs of an API call are validated against the API specification using rules that mirror the validation a real backend would apply. If any check fails, the simulator returns an error message \emph{without} calling the LLM. The checks are:
\begin{itemize}[leftmargin=*,topsep=2pt,itemsep=2pt]
\item \textbf{Unknown parameters} not present in the endpoint's parameter list.
\item \textbf{Missing required parameters} for which no default was filled in.
\item \textbf{Type mismatches}, with the same scalar coercions a real backend would perform (\eg, the string ``42'' is accepted for an integer field, but a boolean is not; container types require an exact match).
\item \textbf{Constraint violations}, including length bounds, numeric bounds (with strict and non-strict inequalities), enumerated value sets, and email-format checks.
\end{itemize}
Surfacing these errors deterministically rather than asking the LLM to invent them ensures that the agent always sees the same error message for the same malformed call, which makes downstream filtering and learning more stable.

\subsection{Prompt Construction}
For every call that reaches the LLM, the simulator builds a two-message conversation: a fixed system prompt that lays out the simulator's role, the six categories of context it receives, and the rules it must follow (\Cref{lst:sim_system_prompt}, with the rules block in
\Cref{lst:sim_rules_a,lst:sim_rules_b}); and a per-call user prompt assembled from the template in \Cref{lst:sim_user_prompt}. The user prompt contains, in this order: the task description that the teacher agent is currently solving; the user profile (first name, last name, email, phone number) drawn from the fictitious supervisor record assigned to this task; the frozen current date and time; the per-app interaction history described in \Cref{app:sim_history}; the fully qualified API name; a JSON dump of the inputs (with defaults filled and the page-limit cap applied); and the full API specification, including the success response schema, which gives the model a clean target for its output. Placing the per-task blocks and the append-only history first means that successive calls within a task share a long common prefix, so only the short tail has to be re-processed. The model is then asked to return raw JSON.

\subsection{Output Schema Validation}
\label{app:sim_schema_check}
Once the LLM's simulated response is received, we first strip any markdown code fences that the model may have emitted in violation of the rules, and then attempt a strict JSON parse. On parse failure, a structured feedback message containing the raw response and the parser error is appended to the conversation (\Cref{app:sim_retry}). After parsing, the JSON value is structurally compared against both (i) the documented success schema and (ii) the canonical error schema $\{\texttt{"message"}: \text{string}\}$. The response is accepted if it conforms to either, since both are legitimate outcomes according to the simulator rules (\Cref{lst:sim_rules_a}). If neither schema matches, both errors are bundled into a structured feedback message, and the loop in \Cref{app:sim_retry} retries.

\subsection{Simulator Judge}
A structurally valid response is still subject to a semantic check, but only for read-only (HTTP \texttt{GET}) APIs. Write APIs (\texttt{POST}, \texttt{PATCH}, \texttt{DELETE}) are accepted as soon as they pass structural validation. This is because their semantic content is typically a simple acknowledgment whose validity is fully captured by the schema, and because gating writes behind a semantic judge would double the LLM cost for state-mutating calls without a proportionate gain in quality. For read APIs, the simulator invokes a separate LLM judge using the system prompt in \Cref{lst:judge_system_prompt} and the user prompt in \Cref{lst:judge_user_prompt}. The judge receives the exact same context as the simulator (specification, inputs, task, history, user profile, and current date/time) alongside the candidate response, and is tasked with returning a JSON verdict containing three fields: a boolean validity flag, a list of issues, and a free-text suggestion block. The judge employs a bounded retry loop to handle any JSON-parsing failures of its own, after which it raises a simulation-failed error.

\subsection{Retry Loop With Structured Feedback}
\label{app:sim_retry}
The structural check and the judge feed into a single retry loop with two independent attempt budgets: a \emph{format} budget for JSON-parse and schema-validation failures, and a \emph{semantic} budget for judge rejections. After any rejection, the conversation is
extended with two messages: an assistant message containing the rejected response verbatim, and a user message containing the structured feedback. For format failures the feedback names the precise error (parse exception, missing keys, type mismatch) and re-states the
two acceptable shapes (success schema or error object). For semantic failures the feedback enumerates the judge's listed issues and reproduces its suggestions verbatim. This in-context conversational format gives the model both a concrete negative example and an actionable hint. We find that it converges within the budget for the vast majority of calls. Once either budget is exhausted, we terminate trajectory synthesis.

\input{tables/simulator_prompt}

\section{Agent}
\label{app:agent}
\input{tables/agent_prompt}

\section{Trajectory Filtering}
\label{sec:app-traj-filtering}
\input{tables/judge_prompt}

\input{sections/7_synthetic_apps}
\ifarxiv\else
\section{Use of AI Assistants}
\label{app:ai-assistants}
We acknowledge the use of AI assistants in the preparation of this manuscript and the implementation of our methodology. Specifically, Gemini-3.1-Pro~\citep{gemini31pro2026} was utilized to proofread and refine the writing of this paper. For software development, we used Claude Code~\citep{claude_code} to assist in implementing our proposed approach. Throughout the development process, we maintained strict oversight by carefully instructing the model and rigorously verifying every piece of generated code to ensure accuracy and correctness. The authors take full responsibility for the integrity of the codebase and the final content of this manuscript.
\fi

%% file: tables/task_throughput.tex
\twotables
{%
    \begin{tabular}{lrr}
    \toprule
    \textbf{Stage} & \textbf{Count} & \textbf{Rate} \\
    \midrule
    Initial generated tasks & 32,009 & 100.00\% \\
    Passed initial judge & 10,509 & 32.83\% \\
    Successfully rewritten & 9,171 & 28.65\% \\
    Passed final judge & 7,072 & 22.09\% \\
    \bottomrule
    \end{tabular}%
}
{Task synthesis throughput for Python coding format \textsc{SimSAT}.}
{tab:syn_task_synthesis_python}
{%
    \begin{tabular}{lrr}
    \toprule
    \textbf{Stage} & \textbf{Count} & \textbf{Rate} \\
    \midrule
    Initial generated tasks & 50,343 & 100.00\% \\
    Passed initial judge & 15,220 & 30.23\% \\
    Successfully rewritten & 13,876 & 27.56\% \\
    Passed final judge & 10,484 & 20.83\% \\
    \bottomrule
    \end{tabular}%
}
{Task synthesis throughput for JSON format \textsc{SimSAT}.}
{tab:syn_task_synthesis_stats_json}

%% file: tables/syn_throughput.tex
\twotables
{%
    \begin{tabular}{lrr}
    \toprule
    \textbf{Stage} & \textbf{Count} & \textbf{Rate} \\
    \midrule
    Input tasks & 10,509 & 100.00\% \\
    Completed trajectories &  8,020 & 76.32\% \\
    Passed judge filtering  & 6,265 & 59.62\% \\
    \midrule
    Simulator API retries & 6,062 / 154,317 & 3.93\% \\
    \bottomrule
    \end{tabular}%
}
{Trajectory synthesis throughput for Python coding format \textsc{SimSAT}.}
{tab:syn_traj_througput_python}
{%
    \begin{tabular}{lrr}
    \toprule
    \textbf{Stage} & \textbf{Count} & \textbf{Rate} \\
    \midrule
    Input tasks & 15,220 & 100.00\% \\
    Completed trajectories & 10,662  & 70.05\% \\
    Passed judge filtering  & 7,241  & 47.58\% \\
    \midrule
    Simulator API retries &  2,454 / 68,436  & 3.59\% \\
    \bottomrule
    \end{tabular}%
}
{Trajectory synthesis throughput for JSON format \textsc{SimSAT}.}
{tab:syn_traj_througput_json}

%% file: tables/syn_failure_distribution.tex
\twotables
{%
    \begin{tabular}{lrr}
    \toprule
    \textbf{Failure Type} & \textbf{Count} & \textbf{Rate} \\
    \midrule
    Judge rejection & 1,755 & 41.35\% \\
    Simulation failure & 1,201 & 28.30\% \\
    Agent's unexpected App use & 936 & 22.05\% \\
    Max solve steps exceeded & 1 & 0.02\% \\
    Agent model execution failure & 350 & 8.25\% \\
    Unexpected & 1 & 0.02\% \\
    \midrule
    Total failures & 4,244 & 100.00\% \\
    \bottomrule
    \end{tabular}%
}
{Distribution of failure modes during trajectory synthesis for Python coding format \textsc{SIMSAT}.}
{tab:syn_failure_distribution}
{%
    \begin{tabular}{lrr}
    \toprule
    \textbf{Failure Type} & \textbf{Count} & \textbf{Rate} \\
    \midrule
    Judge rejection & 3,421 & 42.88\% \\
    Simulation failure & 723 & 9.06\% \\
    Agent's unexpected App use & 1,650 & 20.68\% \\
    Max solve steps exceeded & 2,055 & 25.76\% \\
    Agent model execution failure & 130 & 1.63\% \\
    Unexpected & 0 & 0.00\% \\
    \midrule
    Total failures & 7,979 & 100.00\% \\
    \bottomrule
    \end{tabular}%
}
{Distribution of failure modes during trajectory synthesis for JSON format \textsc{SimSAT}.}
{tab:syn_failure_distribution_json}

%% file: figures/syn_task_dist.tex
\begin{figure*}[t]
    \centering
    \begin{subfigure}{0.24\textwidth}
        \centering
        \includegraphics[width=\linewidth]{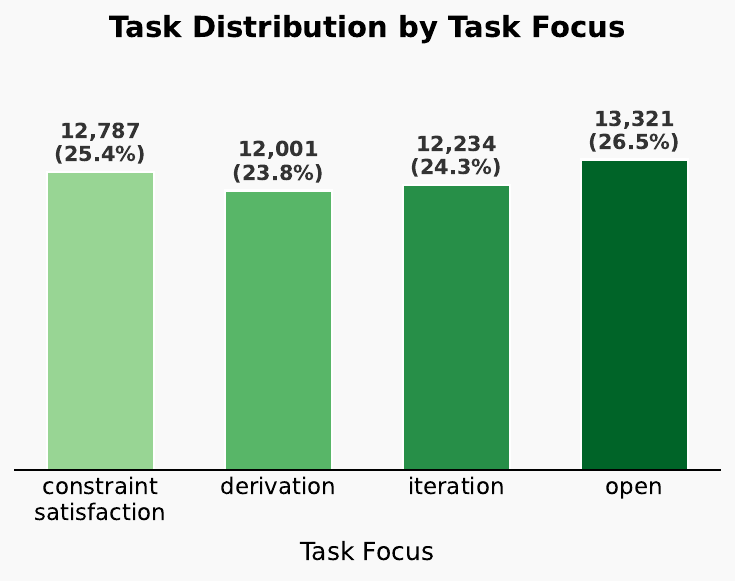}
        \caption{Task focus}
        \label{fig:syn_task_focus}
    \end{subfigure}
    \begin{subfigure}{0.24\textwidth}
        \centering
        \includegraphics[width=\linewidth]{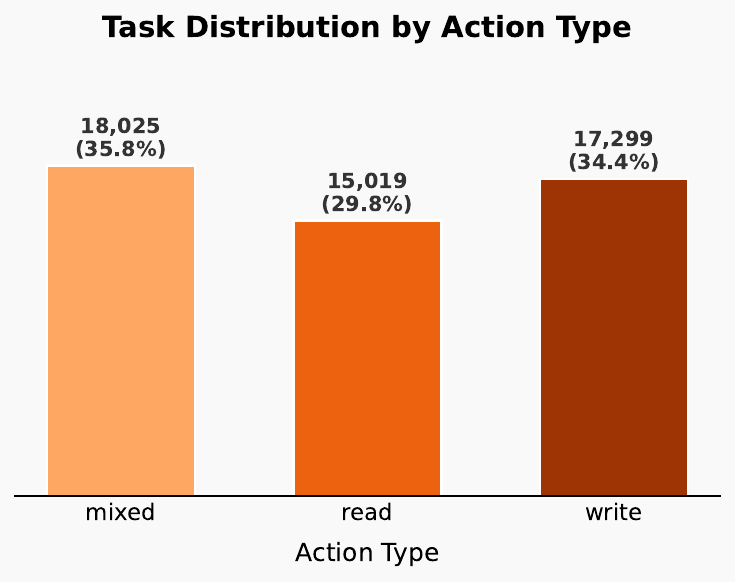}
        \caption{Action type}
        \label{fig:syn_action_type}
    \end{subfigure}
    \begin{subfigure}{0.24\textwidth}
        \centering
        \includegraphics[width=\linewidth]{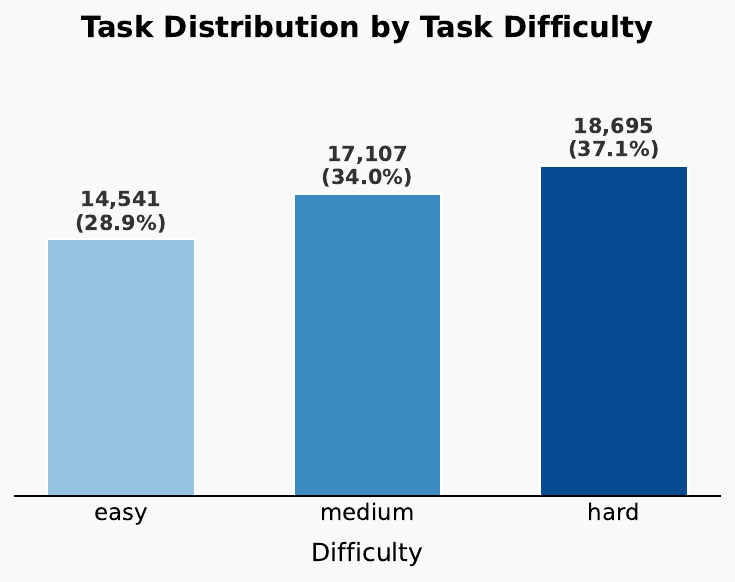}
        \caption{Difficulty}
        \label{fig:syn_difficulty}
    \end{subfigure}
    \begin{subfigure}{0.24\textwidth}
        \centering
        \includegraphics[width=\linewidth]{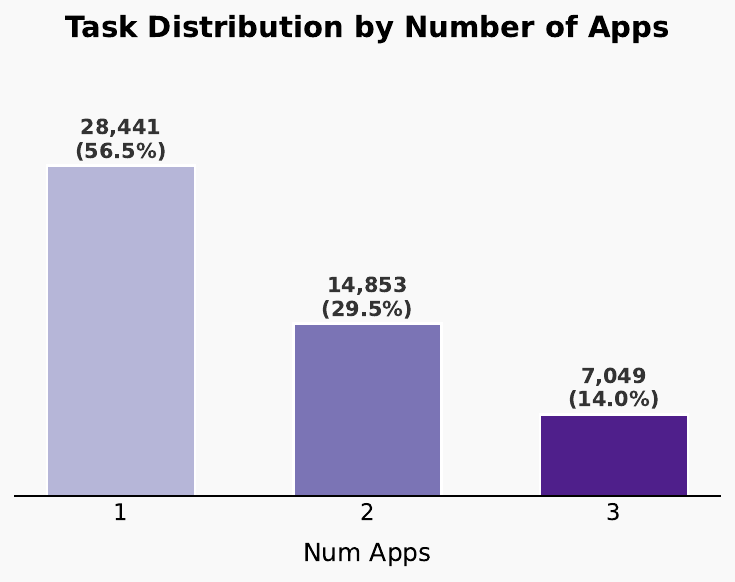}
        \caption{Number of apps}
        \label{fig:syn_num_apps}
    \end{subfigure}

    \begin{subfigure}{0.24\textwidth}
        \centering
        \includegraphics[width=\linewidth]{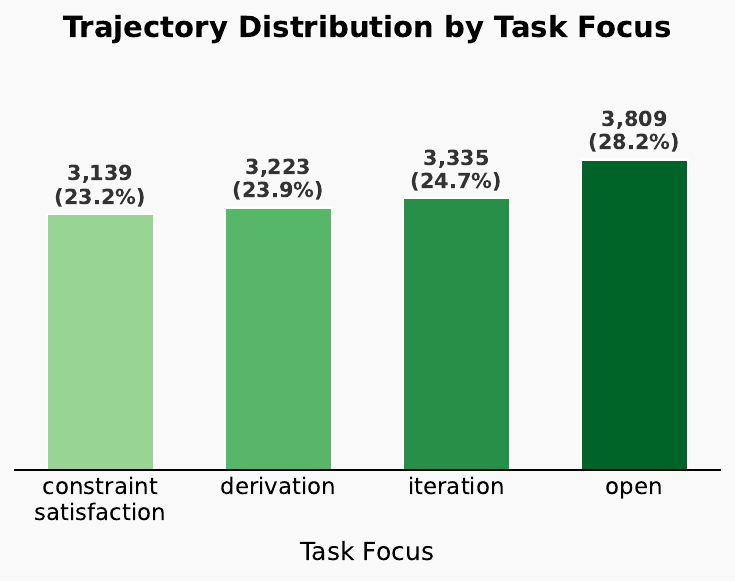}
        \caption{Task focus}
        \label{fig:syn_traj_task_focus}
    \end{subfigure}
    \begin{subfigure}{0.24\textwidth}
        \centering
        \includegraphics[width=\linewidth]{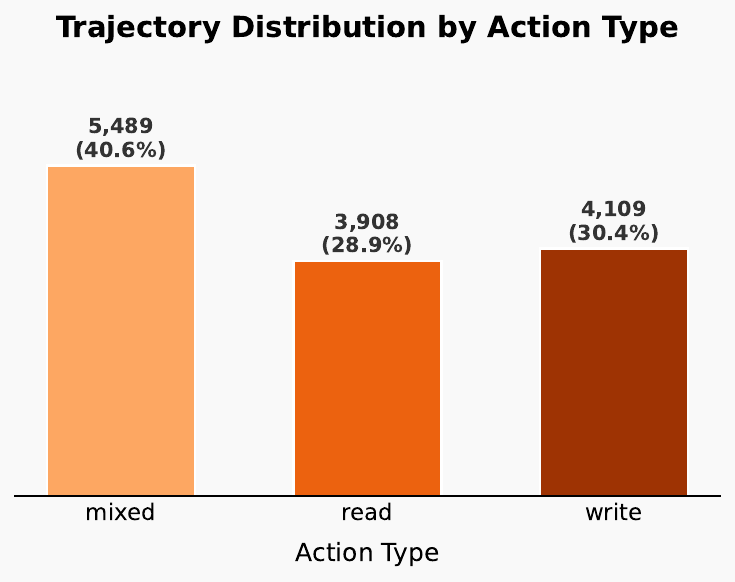}
        \caption{Action type}
        \label{fig:syn_traj_action_type}
    \end{subfigure}
    \begin{subfigure}{0.24\textwidth}
        \centering
        \includegraphics[width=\linewidth]{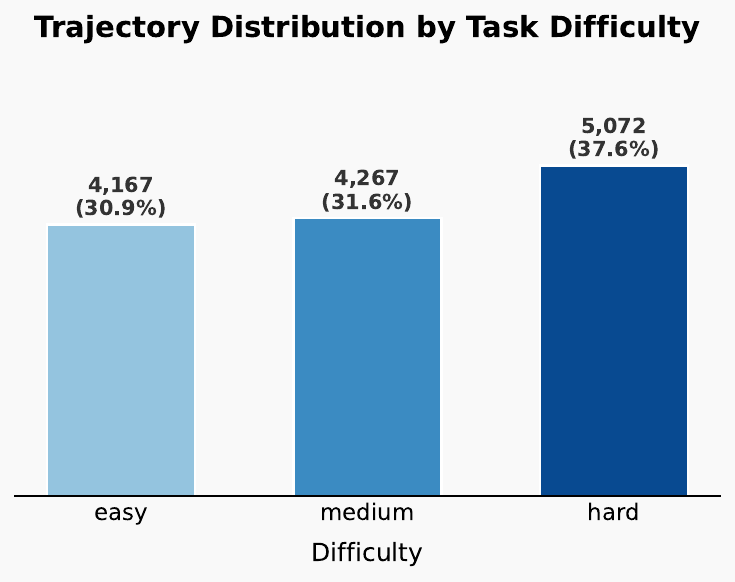}
        \caption{Difficulty}
        \label{fig:syn_traj_difficulty}
    \end{subfigure}
    \begin{subfigure}{0.24\textwidth}
        \centering
        \includegraphics[width=\linewidth]{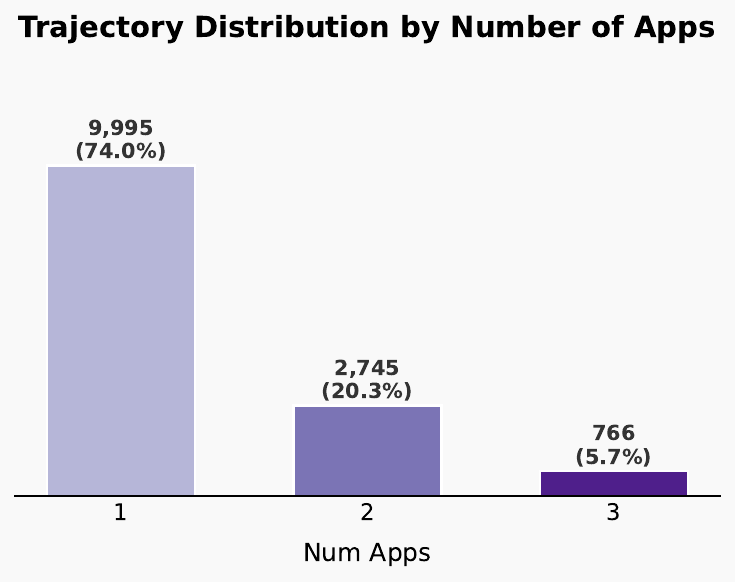}
        \caption{Number of apps}
        \label{fig:syn_traj_num_apps}
    \end{subfigure} 
    \caption{Task and trajectory distributions of the \textsc{SIMSAT} dataset across task focus, action type, difficulty, and the number of apps. The top row shows the generated task pool and  the bottom row shows the filtered trajectories.}
    \label{fig:syn_task_dist}
\end{figure*}

%% file: figures/syn_api_coverage.tex
\begin{figure}[t]
    \centering
    \includegraphics[width=0.95\linewidth]{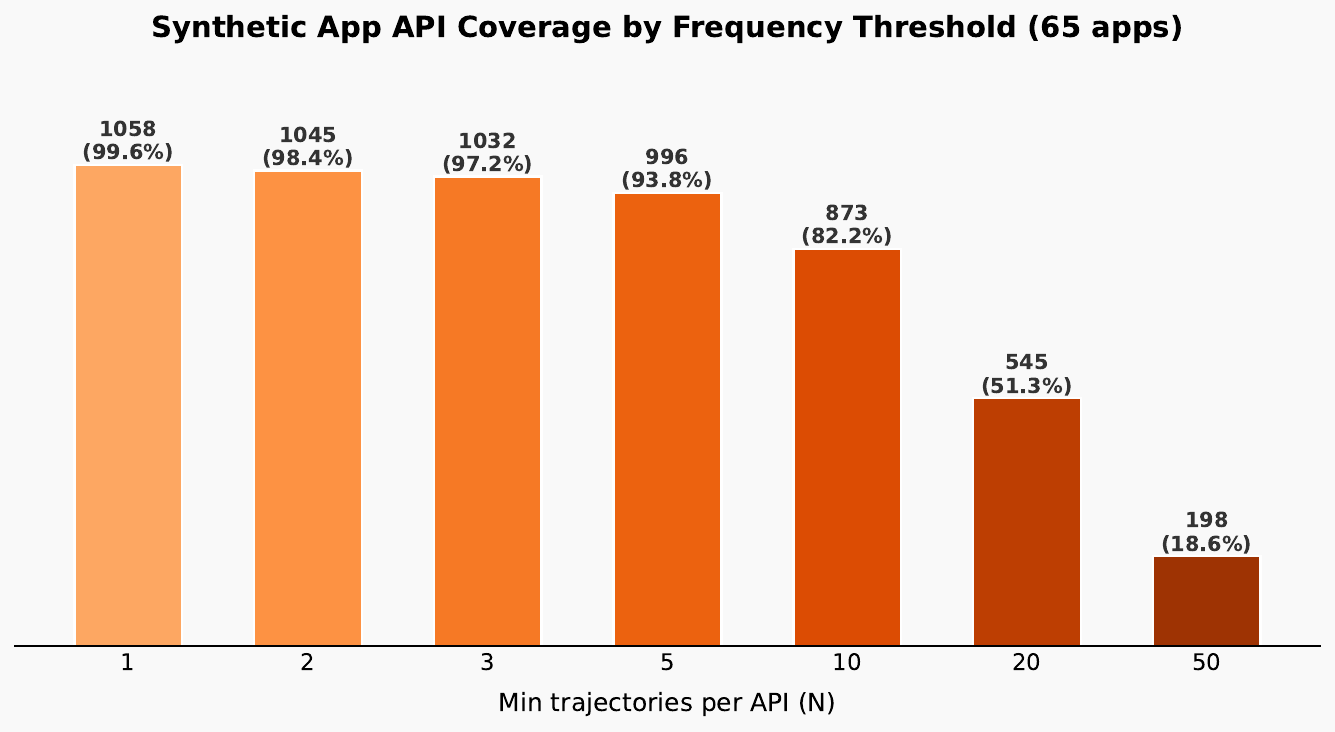}
    \caption{API coverage across our synthetic app APIs, showing the number (and percentage) of APIs called in at least $N$ successful trajectories.}
        \label{fig:syn_api_coverage}
\end{figure}

%% file: tables/appworld_appendix.tex
\begin{table*}[t]
\centering
\resizebox{0.57\linewidth}{!}{
\begin{tabular}[b]{lcccc}
\toprule
 & \multicolumn{2}{c}{\bf{Test-N}} & \multicolumn{2}{c}{\bf{Test-C}} \\
\cmidrule(lr){2-3} \cmidrule(lr){4-5}
 & \bf{TGC} & \bf{SGC} & \bf{TGC} & \bf{SGC} \\
\midrule
\multicolumn{5}{c}{\bf Off-the-shelf models (zero-shot)}\\
\midrule
Gemini-3.1-Pro   & 95.3 $\pm$ 1.1 & 90.2   $\pm$ 2.9 & 86.8   $\pm$ 1.0 & 75.4  $\pm$ 2.1 \\
GLM-5.1-FP8      & 86.5 $\pm$ 2.0 & 77.0   $\pm$ 3.6 & 81.9   $\pm$ 0.8 & 67.5  $\pm$ 1.7 \\
GPT-4o           & 48.8  & 32.1  & 30.2  & 13.0 \\
\midrule

\multicolumn{5}{c}{\bf Qwen3-1.7B}\\
\midrule
zero-shot                   & \phantom{0}0.0 $\pm$ 0.0 & 0.0 $\pm$ 0.0 & \phantom{0}0.1 $\pm$ 0.1 & 0.0 $\pm$ 0.0 \\
\textsc{SIMSAT}           & 13.9 $\pm$ 2.0           & \phantom{0}3.1 $\pm$ 2.3 & \phantom{0}6.3 $\pm$ 1.1 & 0.8 $\pm$ 1.0 \\
\midrule

\multicolumn{5}{c}{\bf Qwen3-4B}\\
\midrule
zero-shot                   & \phantom{0}8.7 $\pm$ 2.2 & 1.8 $\pm$ 1.0 & \phantom{0}3.5 $\pm$ 0.9 & 0.4 $\pm$ 0.5 \\
\textsc{SIMSAT}           & 50.3 $\pm$ 2.5           & 31.5 $\pm$ 3.4 & 37.2 $\pm$ 1.0           & 16.3 $\pm$ 1.5 \\
\midrule

\multicolumn{5}{c}{\bf Qwen3-8B}\\
\midrule
zero-shot                   & 20.2 $\pm$ 2.8 & 5.4 $\pm$ 3.0 & \phantom{0}7.8 $\pm$ 1.5 & 1.7 $\pm$ 0.7 \\
\textsc{SIMSAT}           & 56.6 $\pm$ 3.2 & 35.3 $\pm$ 4.5 & 43.7 $\pm$ 1.9           & 21.8 $\pm$ 2.1 \\
\midrule

\multicolumn{5}{c}{\bf Qwen3-14B}\\
\midrule
zero-shot                   & 33.2 $\pm$ 2.9 & 13.6 $\pm$ 2.3 & 16.3 $\pm$ 1.9 & 5.2 $\pm$ 1.5 \\
\textsc{SIMSAT}           & 66.5 $\pm$ 4.1 & 45.3 $\pm$ 7.3 & 53.7 $\pm$ 1.9 & 29.9 $\pm$ 3.1 \\
\midrule
\multicolumn{5}{c}{\bf Qwen3.5-2B}\\
\midrule
zero-shot                   & \phantom{0}0.1 $\pm$ 0.2 & 0.0 $\pm$ 0.0 & \phantom{0}0.2 $\pm$ 0.3 & 0.0 $\pm$ 0.0 \\
\textsc{SIMSAT}           & 36.0 $\pm$ 2.0           & 17.6 $\pm$ 1.8 & 27.0 $\pm$ 2.1           & 10.3 $\pm$ 2.0 \\
\midrule

\multicolumn{5}{c}{\bf Qwen3.5-4B}\\
\midrule
zero-shot                   & 27.8 $\pm$ 4.5 & 5.8 $\pm$ 3.8 & 17.3 $\pm$ 1.8 & 4.1 $\pm$ 1.3 \\
\textsc{SIMSAT}           & 66.8 $\pm$ 2.0 & 45.1 $\pm$ 3.0 & 58.8 $\pm$ 2.1 & 36.2 $\pm$ 2.2 \\
\midrule

\multicolumn{5}{c}{\bf Qwen3.5-9B}\\
\midrule
zero-shot                   & 25.2 $\pm$ 3.4 & 5.8 $\pm$ 2.7 & 15.4 $\pm$ 1.5 & 3.9 $\pm$ 1.3 \\
\textsc{SIMSAT}           & 74.5 $\pm$ 3.3 & 54.5 $\pm$ 6.8 & 65.4 $\pm$ 1.9 & 44.0 $\pm$ 3.3 \\
\bottomrule
\end{tabular}
}
\caption{Performance on AppWorld test splits with standard deviations across 8 runs. Mean $\pm$ standard deviation is reported. TGC denotes Task Goal Completion and SGC denotes Scenario Goal Completion. AWT denotes AppWorld training data.}
\label{tab:appworld-app}
\end{table*}

%% file: tables/officebench_app.tex
\begin{table*}[t]
\centering
\small
\resizebox{0.58\textwidth}{!}{
\begin{tabular}{lccc}
\toprule
 & \bf{2-app} & \bf{3-app} & \bf{Overall} \\
\midrule
\multicolumn{4}{c}{\bf Off-the-shelf models (zero-shot)}\\
\midrule
Gemini-3.1-Pro & 90.0 $\pm$ 2.7 & 77.7 $\pm$ 4.9 & 83.6 $\pm$ 3.0 \\
GLM-5.1-FP8 & 90.2 $\pm$ 1.5 & 74.3 $\pm$ 2.1 & 82.0 $\pm$ 1.3 \\
GPT-4o & 74.5 & 50.9 & 62.7 \\
\midrule
Qwen3-8B Simia & 53.2 $\pm$ 5.8 & 26.6 $\pm$ 4.0 & 39.4 $\pm$ 3.9 \\
\midrule
\multicolumn{4}{c}{\bf Qwen3-1.7B}\\
\midrule
zero-shot & 22.8 $\pm$ 5.5 & \phantom{0}4.6 $\pm$ 1.4 & 13.3 $\pm$ 2.3 \\
\textsc{SIMSAT} & 36.3 $\pm$ 4.3 & 10.5 $\pm$ 3.6 & 22.9 $\pm$ 2.8 \\
\midrule

\multicolumn{4}{c}{\bf Qwen3-4B}\\
\midrule
zero-shot & 49.8 $\pm$ 4.1 & 38.9 $\pm$ 4.7 & 44.1 $\pm$ 2.4 \\
\textsc{SIMSAT} & 68.9 $\pm$ 3.2 & 51.4 $\pm$ 3.0 & 59.8 $\pm$ 2.6 \\
\midrule

\multicolumn{4}{c}{\bf Qwen3-8B}\\
\midrule
zero-shot & 63.2 $\pm$ 4.2 & 32.7 $\pm$ 3.5 & 47.4 $\pm$ 2.9 \\
\textsc{SIMSAT} & 78.2 $\pm$ 3.2 & 48.4 $\pm$ 4.6 & 62.7 $\pm$ 2.7 \\
\midrule

\multicolumn{4}{c}{\bf Qwen3-14B}\\
\midrule
zero-shot & 69.1 $\pm$ 6.6 & 47.5 $\pm$ 3.4 & 57.9 $\pm$ 3.8 \\
\textsc{SIMSAT} & 78.4 $\pm$ 3.8 & 53.0 $\pm$ 2.3 & 65.2 $\pm$ 1.5 \\
\midrule

\multicolumn{4}{c}{\bf Qwen3.5-2B}\\
\midrule
zero-shot & \phantom{0}9.1 $\pm$ 3.1 & \phantom{0}0.7 $\pm$ 0.9 & \phantom{0}4.7 $\pm$ 1.5 \\
\textsc{SIMSAT} & 45.8 $\pm$ 4.4 & 21.6 $\pm$ 6.0 & 33.3 $\pm$ 4.3 \\
\midrule

\multicolumn{4}{c}{\bf Qwen3.5-4B}\\
\midrule
zero-shot & 62.3 $\pm$ 6.4 & 37.1 $\pm$ 3.9 & 49.2 $\pm$ 3.5 \\
\textsc{SIMSAT} & 79.2 $\pm$ 2.8 & 52.1 $\pm$ 3.2 & 65.1 $\pm$ 1.8 \\
\midrule

\multicolumn{4}{c}{\bf Qwen3.5-9B}\\
\midrule
zero-shot & 68.6 $\pm$ 4.2 & 46.4 $\pm$ 4.4 & 57.1 $\pm$ 2.0 \\
\textsc{SIMSAT} & 79.2 $\pm$ 2.6 & 60.0 $\pm$ 4.0 & 69.2 $\pm$ 2.7 \\
\bottomrule
\end{tabular}
}
\caption{Performance on OfficeBench with standard deviations across 8 runs. Mean $\pm$ standard deviation is reported.}
\label{tab:officebench_std}
\end{table*}

%% file: tables/teacher_appworld.tex
\begin{table*}[t]
\centering
\footnotesize
\setlength{\tabcolsep}{1.5pt}
\begin{tabular*}{\textwidth}{@{\extracolsep{\fill}} l cc cc cc cc cc cc cc @{}}
\toprule
& \multicolumn{2}{c}{\bf Qwen3-1.7B} & \multicolumn{2}{c}{\bf Qwen3-4B} & \multicolumn{2}{c}{\bf Qwen3-8B} & \multicolumn{2}{c}{\bf Qwen3-14B} & \multicolumn{2}{c}{\bf Qwen3.5-2B} & \multicolumn{2}{c}{\bf Qwen3.5-4B} & \multicolumn{2}{c}{\bf Qwen3.5-9B} \\
\cmidrule(lr){2-3} \cmidrule(lr){4-5} \cmidrule(lr){6-7} \cmidrule(lr){8-9} \cmidrule(lr){10-11} \cmidrule(lr){12-13} \cmidrule(lr){14-15}
& Test-N & Test-C & Test-N & Test-C & Test-N & Test-C & Test-N & Test-C & Test-N & Test-C & Test-N & Test-C & Test-N & Test-C \\
\midrule
zero-shot        & \phantom{0}0.0 & \phantom{0}0.1 & \phantom{0}8.7 & \phantom{0}3.5 & 20.2 & \phantom{0}7.8 & 33.2 & 16.3 & \phantom{0}0.1 & \phantom{0}0.2 & 27.8 & 17.3 & 25.2 & 15.4 \\
\midrule
GLM-4.7 teacher  & \bf 14.7 & \bf \phantom{0}6.8 & 45.9 & 29.9 & 53.1 & 34.8 & 62.4 & 45.9 & 32.9 & 22.9 & 59.4 & 47.1 & 68.2 & 54.2 \\
GLM-5.1 teacher  & 13.9 & \phantom{0}6.3 & \bf 50.3 & \bf 37.2 & \bf 56.6 & \bf 43.7 & \bf 66.5 & \bf 53.7 & \bf 36.0 & \bf 27.0 & \bf 66.8 & \bf 58.8 & \bf 74.5 & \bf 65.4 \\
\bottomrule
\end{tabular*}
\vspace{-0.1in}
\caption{AppWorld TGC for models fine-tuned on data generated with different teacher agents. Both conditions use the identical synthesis pipeline and the same GLM-5.1-FP8 API simulator, and differ only in the model acting as the teacher agent; the GLM-5.1 teacher row is the main \textsc{SimSAT} configuration.}
\label{tab:teacher_appworld}
\end{table*}

%% file: tables/teacher_officebench.tex
\begin{table*}[t]
\centering
\footnotesize
\setlength{\tabcolsep}{2pt}
\begin{tabular*}{\textwidth}{@{\extracolsep{\fill}} l cc cc cc cc cc cc cc @{}}
\toprule
& \multicolumn{2}{c}{\bf Qwen3-1.7B} & \multicolumn{2}{c}{\bf Qwen3-4B} & \multicolumn{2}{c}{\bf Qwen3-8B} & \multicolumn{2}{c}{\bf Qwen3-14B} & \multicolumn{2}{c}{\bf Qwen3.5-2B} & \multicolumn{2}{c}{\bf Qwen3.5-4B} & \multicolumn{2}{c}{\bf Qwen3.5-9B} \\
\cmidrule(lr){2-3} \cmidrule(lr){4-5} \cmidrule(lr){6-7} \cmidrule(lr){8-9} \cmidrule(lr){10-11} \cmidrule(lr){12-13} \cmidrule(lr){14-15}
& 2-app & 3-app & 2-app & 3-app & 2-app & 3-app & 2-app & 3-app & 2-app & 3-app & 2-app & 3-app & 2-app & 3-app \\
\midrule
zero-shot        & 22.8 & \phantom{0}4.6 & 49.8 & 38.9 & 63.2 & 32.7 & 69.1 & 47.5 & \phantom{0}9.1 & \phantom{0}0.7 & 62.3 & 37.1 & 68.6 & 46.4 \\
\midrule
GLM-4.7 teacher  & 24.0 & \phantom{0}9.5 & 67.6 & 42.5 & 67.2 & 45.2 & 75.2 & 51.6 & \bf 52.2 & \bf 25.5 & 67.4 & 42.3 & 77.7 & 59.1 \\
GLM-5.1 teacher  & \bf 36.3 & \bf 10.5 & \bf 68.9 & \bf 51.4 & \bf 78.2 & \bf 48.4 & \bf 78.4 & \bf 53.0 & 45.8 & 21.6 & \bf 79.2 & \bf 52.1 & \bf 79.2 & \bf 60.0 \\
\bottomrule
\end{tabular*}
\vspace{-0.1in}
\caption{OfficeBench pass rate for models fine-tuned on data generated with different teacher agents. Both conditions use the identical synthesis pipeline and the same GLM-5.1-FP8 API simulator, and differ only in the model acting as the teacher agent; the GLM-5.1 teacher row is the main \textsc{SimSAT} configuration.}
\label{tab:teacher_officebench}
\end{table*}

%% file: tables/explicit_reasoning.tex
\begin{table}[t]
\centering
\small
\setlength{\tabcolsep}{3pt}
\begin{tabular*}{\narrowtablewidth}{@{\extracolsep{\fill}} l cc cc @{}}
\toprule
& \multicolumn{2}{c}{\bf AppWorld} & \multicolumn{2}{c}{\bf OfficeBench} \\
\cmidrule(lr){2-3} \cmidrule(lr){4-5}
Training data & Test-N & Test-C & 2-app & 3-app \\
\midrule
\textsc{SimSAT}              & \bf 74.5 & \bf 65.4 & \bf 79.2 & \bf 60.0 \\
\quad w/o explicit reasoning & 56.1 & 44.9 & 65.2 & 42.3 \\
\bottomrule
\end{tabular*}
\caption{Effect of removing explicit reasoning from the training trajectories. Qwen3.5-9B is fine-tuned on \textsc{SimSAT} data with the reasoning segments stripped, using the same recipe. We report AppWorld TGC and OfficeBench pass rate.}
\label{tab:explicit_reasoning}
\end{table}

%% file: tables/task_examples.tex
\begin{table*}[t]
\centering
\small
\renewcommand{\arraystretch}{1.25}
\begin{tabular}{p{0.46\linewidth} p{0.46\linewidth}}
\toprule
\multicolumn{2}{l}{\textbf{Example 1.} hard / read / iteration / 1 app \quad\quad Compression: 402 $\rightarrow$ 85 chars (21\%)} \\
\multicolumn{2}{p{0.94\linewidth}}{\textbf{APIs:} $\langle \texttt{expenses\_app} \rangle$\texttt{.show\_no\_group\_expenses}, $\langle \texttt{expenses\_app} \rangle$\texttt{.show\_no\_group\_payments},}\\
\multicolumn{2}{p{0.94\linewidth}}{$\langle \texttt{expenses\_app} \rangle$\texttt{.show\_expense\_comments}, $\langle \texttt{expenses\_app} \rangle$\texttt{.show\_payment\_comments},}\\
\multicolumn{2}{p{0.94\linewidth}}{$\langle \texttt{expenses\_app} \rangle$\texttt{.delete\_payment\_comment}} \\
\cmidrule(lr){1-1} \cmidrule(lr){2-2}
\textbf{Initial task} & \textbf{Rewritten task} \\
\cmidrule(lr){1-1} \cmidrule(lr){2-2}
Help me understand the communication activity on my non-group transactions in the $\langle \texttt{expenses\_app}\rangle$. I want to know how many different unique people have commented on my direct expenses and payments. Go through all expenses and payments that aren't part of any group, check all the comments on each one, and tell me the total count of distinct users who have participated in discussions across all these transactions.
&
How many unique people have commented on my non-group expenses and payments in total on $\langle \texttt{expenses\_app}\rangle?$ \\
\midrule

\multicolumn{2}{l}{\textbf{Example 2.} medium / read / iteration / 1 app \quad\quad Compression: 413 $\rightarrow$ 116 chars (28\%)} \\
\multicolumn{2}{p{0.94\linewidth}}{\textbf{APIs:} $\langle \texttt{money\_transfer\_app} \rangle$\texttt{.show\_notifications}, $\langle \texttt{money\_transfer\_app} \rangle$\texttt{.show\_transaction},}\\
\multicolumn{2}{p{0.94\linewidth}}{$\langle \texttt{money\_transfer\_app} \rangle$\texttt{.show\_transaction\_comments}, $\langle \texttt{money\_transfer\_app} \rangle$\texttt{.show\_account},}\\
\multicolumn{2}{p{0.94\linewidth}}{$\langle \texttt{money\_transfer\_app} \rangle$\texttt{.show\_profile}} \\
\cmidrule(lr){1-1} \cmidrule(lr){2-2}
\textbf{Initial task} & \textbf{Rewritten task} \\
\cmidrule(lr){1-1} \cmidrule(lr){2-2}
I have unread notifications in $\langle \texttt{money\_transfer\_app} \rangle$ and I want to know what's going on with them. Go through all my unread notifications, find any that are related to transactions, and check those transactions to see if people have left comments. For each comment you find, look up the person's profile to see who they are. Give me a summary of which notifications have related activity in comments and what people are saying in those comments.
&
Check my unread $\langle \texttt{money\_transfer\_app} \rangle$ notifications for any transactions with comments and tell me who commented and what they said. \\

\bottomrule
\end{tabular}
\caption{Examples of task rewriting in \textsc{SIMSAT}. Each row shows an initial generated task (left) and its rewritten counterpart (right). Configuration tags indicate \emph{difficulty / action type / task focus / number of apps}. Rewriting consistently compresses verbose, step-by-step instructions into concise, intent-level user requests.}
\label{tab:rewrite_examples}
\end{table*}

%% file: tables/task_prompt.tex
%

\subsection{Prompts for Task Synthesis}
\paragraph{Task generation prompt.}
\Cref{lst:taskgen_prompt} shows the prompt used by the task generator. Given the documentation of a sampled subset of APIs, a target difficulty / action-type / task-focus configuration, and a list of negative examples (already-generated tasks) and under-utilized APIs, the model produces \texttt{\{num\_tasks\}} candidate tasks per call. The configurable clauses (\texttt{\{difficulty\_clause\}}, \texttt{\{action\_type\_clause\}}, \texttt{\{task\_focus\_clause\}}) are sampled from the variants enumerated in \Cref{tab:taskgen_difficulty,tab:taskgen_action,tab:taskgen_focus}.

\begin{figure*}[t]
\begin{lstlisting}[style=prettyprompt, captionpos=b, caption={Task generation prompt.}, label={lst:taskgen_prompt}]
You are an AI model with API understanding capabilities. You are provided with documentation of various APIs supported by commonly used mobile phone apps within <tools></tools> XML tags below:
<tools>
{api_docs_context}
</tools>

Your job is to generate realistic tasks that users would ask a powerful coding-based tool calling agent to solve by utilizing the provided APIs. The agent is allowed to solve the task in an iterative multi-step fashion. In each step, the agent can produce python code (including if/else and loops if needed) that invokes one or multiple APIs. This python code will be executed using an environment with API execution support and a backend user database, and the execution results will be provided to the agent before the next step.

Carefully go through all the provided API documentations which include the input parameter descriptions and the output response schemas for each API. Then, generate {num_tasks} distinct tasks each satisfying all the below requirements:

# Task Requirements:
- **Solvability:** The task should be completely solvable by an agent using the provided APIs without any additional information from the user. You can assume that the agent already knows the user's name, username, email, phone number, birthday, sex, address, payment cards, login access tokens for all the apps, and current date/time.
- **Specificity:** Use specific details (names, amounts, dates etc) only where needed to make the task unambiguous. Do not include anything the agent can look up using the APIs.
- **Clarity:** The task should not have contradicting or confusing statements.
- **Interdependent subtasks:** The task SHOULD NOT be a simple concatenation of completely independent subtasks. Solving the task should require the agent to solve many interdependent subtasks.
- **Referential complexity** When possible, introduce indirect references to entities (people, objects, locations, amounts, etc.) as long as these references can be resolved completely using the given APIs.
- **Realism:** Write the task exactly as a real user would say it to a personal assistant — natural, high-level, and goal-oriented. The task MUST NOT read like step-by-step instructions. The task should only state what the user wants and the agent will figure out how to achieve it.
- **Coverage:** Each task should cover {min_apis_per_app}-{max_apis_per_app} apis from each app.
- **Required apps:** Every task MUST use at least one of the following apps: {required_apps}. Tasks that do not involve any of these apps are invalid and will be rejected.The APIs covered by each task should vary between the tasks.
- **Difficulty:** {difficulty_clause}
- **Action type:** {action_type_clause}
- **Task focus:** {task_focus_clause}

{focus_section}
{negative_section}
First carefully think about how to construct tasks that satisfy all the above requirements. Specifically, think about which subsets of the provided APIs can result in meaningful tasks and identify several such subsets to generate multiple tasks. Do not make any assumptions about any API. The API documentation clearly describes the input parameters and the expected output. DO NOT hallucinate any new APIs that are not part of the APIs provided within <tools></tools> tags. For each task, first generate the names of all the APIs (comma separated) covered by the task in a separate line starting with `APIs COVERED:`, then describe how an agent can fully solve the task using these APIs in a separate line starting with `EXPLANATION:`, and finally generate the task in a separate line that starts with `TASK:`. You should not include any additional information in this line other than the task itself. DO NOT generate any additional summaries or titles for the tasks.
\end{lstlisting}
\end{figure*}

\begin{table*}[t]
\centering
\small
\renewcommand{\arraystretch}{1.3}
\begin{tabular}{l p{0.85\linewidth}}
\toprule
\textbf{Level} & \textbf{Clause} \\
\midrule
\textsc{easy}   & The task requires the agent to perform a focused, self-contained operation. The solution would be a linear sequence of API calls with minimal branching. Filtering, sorting, or iterating over items would be the main challenge. \\
\midrule
\textsc{medium} & The task introduces indirect references that the agent must resolve via API calls before acting (e.g.\ ``the person I owe the most'', ``my most recent group'', ``the highest-rated item''). The output of one subtask feeds directly into the next, creating a clear dependency chain. The solution requires moderate branching or iteration. \\
\midrule
\textsc{hard}   & The task requires multi-hop reasoning: the agent must gather context across multiple apps, reconcile entities, and execute a coordinated sequence of actions. For example, an entity resolved in app A feeds a lookup in app B which feeds an action in app C. The solution requires conditional branches or loops over dynamic lists. \\
\bottomrule
\end{tabular}
\caption{Difficulty clauses inserted into the task generation prompt. Exactly one is sampled per task.}
\label{tab:taskgen_difficulty}
\end{table*}

\begin{table*}[t]
\centering
\small
\renewcommand{\arraystretch}{1.3}
\begin{tabular}{l p{0.85\linewidth}}
\toprule
\textbf{Type} & \textbf{Clause} \\
\midrule
\textsc{mixed} & The task should be a MIXED task that requires substantial reading AND writing: the agent must first gather and reason over information from one or more apps (\texttt{GET} requests), then use that information to drive meaningful write operations (\texttt{POST}, \texttt{PATCH}, \texttt{DELETE}) in the same or different apps. Neither reads nor writes dominate --- both are essential to completing the task. \\
\midrule
\textsc{read}  & The task should be a READ-HEAVY task. The task primarily requires the agent to query, search, look up, and report information. The majority of API calls should be \texttt{GET} requests (lookup, list, search, show). Write operations should be minimal or absent --- the output of the task is a piece of information or a summary, not a state change in any app. \\
\midrule
\textsc{write} & The task should be a WRITE-HEAVY task. The task primarily requires the agent to create, update, send, buy, delete, or modify data. The majority of API calls should be \texttt{POST}, \texttt{PATCH}, or \texttt{DELETE} requests. Read operations are only used minimally to look up IDs or verify preconditions before writing. \\
\bottomrule
\end{tabular}
\caption{Action-type clauses inserted into the task generation prompt. Exactly one is sampled per task.}
\label{tab:taskgen_action}
\end{table*}

\begin{table*}[t]
\centering
\small
\renewcommand{\arraystretch}{1.3}
\resizebox{1\linewidth}{!}{%
\begin{tabular}{l p{0.82\linewidth}}
\toprule
\textbf{Focus} & \textbf{Clause} \\
\midrule
\textsc{constraint\ satisfaction} & The task should require the agent to find or select entities that satisfy MULTIPLE simultaneous criteria. The agent must filter, search, or verify that several conditions hold at the same time before acting. Examples of constraints: price within a range,  rating above a threshold, date before a deadline, belonging to a specific category, matching a text pattern, being associated with a particular person. The task should require at least 2--3 distinct constraints to be checked together. The challenge is in correctly combining the constraints, not in the action itself. \\
\midrule
\textsc{derivation} & The task should require the agent to derive a SINGLE piece of new information by aggregating or comparing multiple data points, then act on that derived result. The derived value is essential --- the agent cannot complete the task without first computing it. Examples: compute total spending to decide a budget, find who owes the most and message that one person, calculate average rating to set a threshold, determine which playlist has the longest total duration and play it.
The task must involve an aggregation step (sum, average, max, min, count, ranking, or comparison) whose SINGLE result drives a subsequent action. The final action should target one entity or a fixed number of entities --- NOT loop over every item in a set. \\
\midrule
\textsc{iteration} & The task should require the agent to perform an operation on EACH item in a set whose membership is discovered at runtime. The set is not known in advance --- the agent must first query to find which items qualify, then process each one individually. Each iteration may involve multiple API calls. Examples: for each person who owes me  money send a reminder, for every overdue task add a comment, for all notes tagged `urgent' share them with a colleague, for each expense above \$50 update its description. The task should NOT be solvable with a single bulk API call --- individual per-item processing is required. \\
\midrule
\textsc{open} & \emph{(no additional focus constraint)} \\
\bottomrule
\end{tabular}%
}
\caption{Task-focus clauses inserted into the task generation prompt. Exactly one is sampled per task. The \textsc{open} setting adds no additional constraint.}
\label{tab:taskgen_focus}
\end{table*}

\paragraph{Task judge prompt.}
After a candidate task is generated, a judge LLM verifies that it satisfies the same requirements that were given to the generator. \Cref{lst:judge_prompt} shows the full prompt template.

\begin{figure*}[t]
\begin{lstlisting}[style=prettyprompt, captionpos=b, caption={Task judge prompt.}, label={lst:judge_prompt}]
Your job is to evaluate the quality of agentic tasks generated by a task generator. The task generator creates tasks based on the descriptions of various APIs provided to it.

Here are the API descriptions provided to the task generator:
{api_docs_context}

Here are the task requirements given to the task generator:
{task_requirements}
Evaluate if the task satisfies all the requirements. In order to evaluate, you should briefly think about how to solve the task. This will help you to figure out the API coverage, the task difficultly and its focus. DO NOT THINK FOR TOO LONG. First provide a brief reasoning behind your decision and then provide your final YES/NO answer in a separate line that starts with `ANSWER:`. You should not include any additional information in this line other than the YES/NO answer itself.

Here is the task to evaluate:
{task_description}
\end{lstlisting}
\end{figure*}

\paragraph{Task rewriting prompt.}
Tasks that pass the judge are then rewritten to remove procedural scaffolding and sound like a natural user request. \Cref{lst:rewrite_prompt} shows the prompt body, and \Cref{lst:rewrite_examples} shows the three in-context examples appended after it.

\begin{figure*}[t]
\begin{lstlisting}[style=prettyprompt, captionpos=b, caption={Task rewriting prompt (body).}, label={lst:rewrite_prompt}]
You are an AI model with API understanding capabilities. Your job is to generate realistic tasks that users would ask a powerful coding-based tool calling agent to solve by utilizing a set of available APIs. The agent is allowed to solve the task in an iterative multi-step fashion. In each step, the agent can produce python code (including if/else and loops if needed) that invokes one or multiple APIs. This python code will be executed using an environment with API execution support and a backend user database, and the execution results will be provided to the agent before the next step.

You will be provided with a task and the documentations of APIs that are needed for solving this task. Carefully go through all the provided API documentations which include the input parameter descriptions and the output response schemas for each API. Then, rewrite the given task in a few sentences such that the rewritten task satisfies all the below requirements:

# Task Requirements:
{task_requirements}
Make sure that the rewritten task also requires all the APIs needed for solving the original task. Do not add any new information or change the intent/scope of the task. Make sure to remove redundant setup phrases, obvious intermediate steps that any agent would infer, and overly explicit instruction-style wording that describes the procedure rather than the goal. Make sure that you do not remove any details that are crucial to what the user exactly wants. Examples of things that must be kept:
    * Specific file names, folder paths, or formats (e.g. 'save it as a .txt file', 'store in ~/reports/', 'delete the zip file after')
    * Specific actions the user explicitly wants (e.g. 'then delete it', 'send them a message', 'create a note')
    * Specific names, amounts, dates, or identifiers
    * Any condition or constraint the user cares about (e.g. 'only if they haven't texted me')

First provide a brief explanation of your rewriting process. Then provide the list of all the APIs (comma separated) covered by the rewritten task in a separate line starting with `APIs COVERED:`, and finally provide the rewritten task in a separate line that starts with `TASK:`. You should not include any additional information in this line other than the task itself.

Here is the task to rewrite:
{task_description}

Here are the APIs required by this task:
{api_docs_context}
\end{lstlisting}
\end{figure*}

\begin{figure*}[t]
\begin{lstlisting}[style=prettyprompt, captionpos=b, caption={Task rewriting prompt (in-context examples appended to the body).}, label={lst:rewrite_examples}]
Here are some in-context examples to guide you with the rewriting process:
## Example 1:
Original: I want to organize sharing photos with my family. Find all my family contacts and check whether I have created a dedicated folder for each person named `<first_name>_<last_name>` in `~/photos/`. If they do have a folder, determine whether we have ever exchanged text messages. For any family member who has a photo folder but whom I have never texted, create a note titled `<first_name> <last_name> Photos` to remind me that I need to share photos with them.

Rewritten: For any of my family contacts who have a folder named `<first_name>_<last_name>` in `~/photos/` but whom I have never texted, create a note titled `<first_name> <last_name> Photos` stating that I need to share photos with them.

## Example 2:
Original: I need to check up on the people listed in my `Project Alpha Team` note. They are supposed to submit their reports as text files named `<first_name>_<last_name>.txt` in the `~/project_alpha/reports/` folder. Determine who hasn't submitted their report yet. For those whose reports are missing, check whether they have sent me any text messages recently. Provide a list of the team members who are both missing their reports and haven't been communicating via text.

Rewritten: Who are the team members listed in my `Project Alpha Team` note who haven't submitted their reports named `<first_name>_<last_name>.txt` to the `~/project_alpha/reports/` folder and also haven't texted me recently?

## Example 3:
Original: Download the latest invoice from my email, save it to ~/invoices/latest.pdf, and then delete the email.

Rewritten: Save the latest invoice from my email to ~/invoices/latest.pdf, then delete that email.
\end{lstlisting}
\end{figure*}

%% file: tables/simulator_prompt.tex
%
%

\subsection{Prompt for Simulator}
\label{sec:simulator-prompt}
\paragraph{Simulation rules.}
\Cref{lst:sim_rules_a,lst:sim_rules_b} list the simulation rules that the simulator must follow when generating an API response, and that the judge uses to verify the simulator's output. They are inlined verbatim into both the simulator and the judge system prompts (\Cref{lst:sim_system_prompt,lst:judge_system_prompt}).

\begin{figure*}[t]
\begin{lstlisting}[style=prettyprompt, captionpos=b, caption={Simulation rules (1/2): user data, response schema, and list generation.}, label={lst:sim_rules_a}]
1. USER-SPECIFIC DATA:
   - Treat the provided user as the currently logged-in user.
   - When simulating user information such as phone, email, address, etc. use the user profile information provided to you.
   - When simulating APIs that return user-specific data (e.g., show_playlist_library, show_orders, show_payment_history), you should simulate data that is aligned with the provided user profile.

2. RESPONSE SCHEMA:
    - For valid API calls, return valid JSON output matching the exact response schema provided.
        - Match all field names, types, and nesting exactly as specified
        - DO NOT copy the example values from the schema; generate your own realistic values instead
    - Return JSON output in the form of {"message": "error description"} instead of the provided response schema if
        - Some required input parameters are missing
        - Some input parameters are invalid (wrong type, violate constraints, format violations, etc.)
        - The API call is clearly infeasible according to the previous API calls (for example deleting/updating/accessing items that don't exist according to the previous API calls)
   - Only generate raw JSON response, no additional text or markdown code blocks

3. GENERATING LIST/ARRAY:
    - Decide the number of items to generate based on the API documentation, input parameter values, and previous API call history.
    - You should generate the exact number of items specified by the `page_limit` parameter in most of the cases. However, if generating that many items is not realistic or meaningful within the context of the API you are simulating, it is acceptable to return fewer items. For example, if simulating reviews for a song written by a user, returning only 1 review is okay since a single user will typically write only one review for a song.
    - If `page_index` value is greater than 1, return empty list unless the previous API call history indicates the presence of items beyond what has been returned so far for this API. For example, an album had 15 songs to begin with according to a previous simulated API call. Then 10 more songs were explicitly added by subsequent API calls. When retrieving the songs in that album, there should be a total of 25 songs. If 10 songs were returned with `page_index=0` and 10 more were returned with `page_index=1`, then you should return the remaining songs with `page_index=2`.
\end{lstlisting}
\end{figure*}

\begin{figure*}[t]
\begin{lstlisting}[style=prettyprompt, captionpos=b, caption={Simulation rules (2/2): data consistency, realism, and diversity.}, label={lst:sim_rules_b}]
4. DATA CONSISTENCY:
   - When current API call refers to some underlying data that a previous API call also refers to, make sure that the current simulation result is consistent with the previous simulation result (e.g., if a song was simulated with a particular artist name in a previous call, make sure that song appears when simulating a list of songs for the same artist in a later call)
   - Sometimes previously simulated data may be modified by a later API call. If current API call is referring to the modified data, make sure that the current simulated data preserves the modifications.
   - Keep aggregate fields such as counts, ratings, and totals consistent with the operations in the history (e.g., like -> like_count increases; unlike -> like_count decreases)
   - Make sure you are not using entities that have been explicitly deleted by previous API calls.

5. DATA REALISM AND CONSTRAINTS:
   - Uniqueness: When generating multiple entities, make sure they have unique identifiers.
   - Timestamps: All timestamps must be relative to the current date/time provided:
     * Past events (orders, messages, created_at) should have timestamps before the current time
     * Future events (appointments, scheduled tasks, reminders) should have timestamps after the current time
     * Recent activity should be within the last few days/weeks from current time
     * Use chronologically ordered timestamps (created_at before updated_at)
   - Numeric ranges: Respect field constraints (e.g., ratings 0-5, counts >= 0, durations > 0)
   - Realistic values: Generate plausible data (real-sounding names, valid email formats, reasonable dates, prices, etc.)

6. DATA DIVERSITY:
   - DO NOT generate only task-relevant entities. Include a mix of relevant and irrelevant items to simulate real-world data.
   - Example: If the task the agent is solving is "find song X by artist Y", a search for "X" should return multiple songs named "X" by different artists, not just the exact match.
   - Aim for roughly 30-50% task-relevant items in list results, with the rest being plausible but not exactly related to the user task.
\end{lstlisting}
\end{figure*}

\paragraph{Simulator system prompt.}
\Cref{lst:sim_system_prompt} shows the system prompt given to the simulator. The placeholder \texttt{\{simulation\_rules\}} is replaced with the rules block in \Cref{lst:sim_rules_a,lst:sim_rules_b} at runtime.

\begin{figure*}[t]
\begin{lstlisting}[style=prettyprompt, captionpos=b, caption={Simulator system prompt.}, label={lst:sim_system_prompt}]
You are an API simulation engine that simulates APIs of Apps installed on a mobile phone.

Your job is to simulate realistic responses for API calls made by an agent that is trying to solve a specific task given to it by the user of the device.
You will be provided with the following information:
1. API documentation (description of API, input parameter descriptions, response schema)
2. Task being solved by the agent
3. History of previous API calls and their simulated results (to maintain consistency)
4. Mobile phone user information
5. Current date and time
6. API call input parameter values

CRITICAL RULES:

{simulation_rules}
\end{lstlisting}
\end{figure*}

\paragraph{Simulator user prompt template.}
For each API call issued by the agent, the simulator is also given a per-call user message that bundles the task context, the user profile, the current date and time, the per-app interaction history, and finally the API specification and call inputs. \Cref{lst:sim_user_prompt} shows the template. The block order is deliberate: the per-task blocks and the append-only history come first, so that consecutive calls within a task share a long common prefix and only the short tail has to be re-processed.

\begin{figure*}[t]
\begin{lstlisting}[style=prettyprompt, captionpos=b, caption={Simulator user prompt template (one per simulated API call).}, label={lst:sim_user_prompt}]
Generate a response to the given API call based on the following context:

## Task Context
The API call is made by an agent that is solving the following user task:
{task_instruction}

## User Profile
First Name: {first_name}
Last Name: {last_name}
Email: {email}
Phone Number: {phone_number}

## Current Date and Time
Date: {date}
Time: {time}

## Previous API Call History
{history_context}

## API Name
{app_name}.{api_name}

## API Call Input Parameters
```json
{inputs}
```

## API Documentation
```json
{api_doc}
```

Generate a JSON response with realistic data that is consistent with the provided parameter values, previous API calls/responses in the history, the task context, the user profile and the current date/time.

Output ONLY the JSON response, nothing else.
\end{lstlisting}
\end{figure*}

\paragraph{Simulator-judge system prompt.}
\Cref{lst:judge_system_prompt} shows the system prompt given to the judge. The judge sees the same rules block as the simulator (\Cref{lst:sim_rules_a,lst:sim_rules_b}) and is asked to evaluate whether the simulator's output complies with each rule.

\begin{figure*}[t]
\begin{lstlisting}[style=prettyprompt, captionpos=b, caption={Judge system prompt.}, label={lst:judge_system_prompt}]
You are a judge evaluating the validity of the response provided by an API simulation engine.

The simulator's job is to simulate realistic responses for API calls made by an agent that is trying to solve a specific task given to it by the user of a mobile phone.

You will be provided with the following information which is also provided to the simulator:
1. API documentation (description of API, input parameter descriptions, response schema)
2. Task being solved by the agent
3. History of previous API calls and their simulated results (to maintain consistency)
4. Mobile phone user information
5. Current date and time
6. API call input parameter values

Here are the set of rules that the simulator is expected to follow:

{simulation_rules}

Your job is to evaluate whether the response provided by the API simulator follows all the rules.
Note that when generating lists, the number of entries returned by the simulator depends on the `page_limit` value provided in the API call. So, even when referring to data that was already simulated in previous API calls, the simulator may not give out all the data at once due to `page_limit` value.
\end{lstlisting}
\end{figure*}

\paragraph{Simulator-judge user prompt template.}
For each candidate simulator response, the judge is given a per-call user message that mirrors the simulator's block order and appends the response under evaluation. The judge returns a JSON verdict with a validity flag, a list of issues, and suggestions for fixing them. \Cref{lst:judge_user_prompt} shows the template.

\begin{figure*}[t]
\begin{lstlisting}[style=prettyprompt, captionpos=b, caption={Judge user prompt template (one per simulated API call).}, label={lst:judge_user_prompt}]
Validate the simulated response of the given API call based on the following context:

## Task Context
The API call is made by an agent that is solving the following user task:
{task_instruction}

## User Profile
First Name: {first_name}
Last Name: {last_name}
Email: {email}
Phone Number: {phone_number}

## Current Date and Time
Date: {date}
Time: {time}

## Previous API Call History
{history_context}

## Simulated API Name
{app_name}.{api_name}

## Simulated API Call Input Parameters
```json
{inputs}
```

## Simulated API Documentation
```json
{api_doc}
```

## Simulation Response to Evaluate
```json
{result}
```

Return a JSON object with exactly the following fields:
{
    "is_valid": true or false,
    "issues": ["<issue 1>", "<issue 2>", ...],
    "suggestions": "<specific suggestions for fixing the issues>"
}

- "issues" should be an empty list if is_valid is True
- "suggestions" should be an empty string if is_valid is True
- Output ONLY the JSON object, no other text.
\end{lstlisting}
\end{figure*}

%% file: tables/agent_prompt.tex
%
%
%

\paragraph{AppWorld agent prompt.}
\Cref{lst:agent_prompt_intro,lst:agent_prompt_instructions} show the full prompt used by the AppWorld ReAct agent. The prompt is a single \texttt{USER} message parsed into OpenAI format. \Cref{lst:agent_prompt_intro} contains the role intro, the three core APIs for exploring documentation, and the REPL loop explanation. \Cref{lst:agent_prompt_instructions} contains the key-instructions block and the runtime task injection point. Placeholders (\texttt{\{\{ supervisor.first\_name \}\}}, \texttt{\{\{ instruction \}\}}, etc.) are filled in at runtime.

\paragraph{OfficeBench agent prompt.}
\Cref{lst:officebench_prompt_system,lst:officebench_prompt_steps} show the prompt used by the OfficeBench agent, loaded from \texttt{configs/prompts\_v2.json}. \Cref{lst:officebench_prompt_system} contains the chat system message: the current-date preamble, the operating rules, the safety constraint, the \texttt{<think>}/\texttt{<answer>} response format, a one-shot example interaction, and the runtime list of installed apps. \Cref{lst:officebench_prompt_steps} contains the three per-step user prompts that are appended at each turn: \texttt{prompt\_undecided\_app} (no app active), \texttt{prompt\_undecided\_app\_w\_history} (no app active, prior steps exist), and \texttt{prompt\_decided\_app} (an app is currently active). Placeholders (\texttt{\{task\}}, \texttt{\{available\_apps\}}, \texttt{\{app\_introduction\}}, \texttt{\{detailed\_instruction\}}, etc.) are filled in at runtime.

\begin{figure*}[t]
\begin{lstlisting}[style=prettyprompt, captionpos=b, caption={Agent prompt (1/2): role intro, the three core documentation APIs, and the REPL execution loop.}, label={lst:agent_prompt_intro}]
USER:
I am your supervisor and you are a super intelligent AI Assistant whose job is to achieve my day-to-day tasks completely autonomously.

To do this, you will need to interact with app/s using their associated APIs on my behalf. For this you will undertake a *multi-step conversation* using a python REPL environment. That is, you will write the python code and the environment will execute it and show you the result, based on which, you will write python code for the next step and so on, until you've achieved the goal. This environment will let you interact with app/s using their associated APIs on my behalf.

Here are three key APIs that you need to know to get more information

# To get a list of apps that are available to you.

```python
print(apis.api_docs.show_app_descriptions())

To get the list of apis under any app listed above, e.g. XYZ

print(apis.api_docs.show_api_descriptions(app_name='XYZ'))

To get the specification of a particular api, e.g. XYZ app's login api

print(apis.api_docs.show_api_doc(app_name='XYZ', api_name='login'))

Each code execution will produce an output that you can use in subsequent calls. Variables defined in a step are accessible in every subsequent step. So, use variables to store any information that may be useful in the following steps. Using these APIs, you can now generate code, that I will execute, to solve the task.
\end{lstlisting}
\end{figure*}

\begin{figure*}[t]
\begin{lstlisting}[style=prettyprompt, captionpos=b, caption={Agent prompt (2/2): the key-instructions block followed by the runtime task injection point. \texttt{{{ instruction }}} is replaced with the actual task description at inference time.}, label={lst:agent_prompt_instructions}]
Key instructions:
(1) Make sure to start code blocks with python followed by a newline(\n) and end code blocks with  followed by a newline(\n).
(2) If you want to call a particular api, then you must refer to it using the format apis.<app_name>.<api_name>
(3) If an app has a "login" api, then you must first log in to the app using this api before using any other api from that app. You can find login credentials using "print(apis.supervisor.show_account_passwords())"
(4) You can use the APIs in the "supervisor" app to get additional information about me and use the APIs in the "phone" app to get information about friends and family.
(5) Always look at API specifications (using apis.api_docs.show_api_doc) before calling an API. When retrieving app/api documentation using api_docs app, do not query for multiple documentations in a single step.
(6) Write only one chunk of code in every step. You can write a multi-line code chunk that makes multiple API calls within one step as long as you have all the information needed to make these calls. You can implement complex logic within a single step when needed using python if/else branching, for/while loops, etc. Make sure everything is working correctly before making any irreversible change.
(7) Many APIs return items in "pages". Make sure to run through all the pages by looping over page_index.
(8) Once you have completed the task, make sure to call apis.supervisor.complete_task(). If the task asked for some information, return it as the answer argument, i.e. call apis.supervisor.complete_task(answer=<answer>). Many tasks do not require an answer, so in those cases, just call apis.supervisor.complete_task() i.e. do not pass any argument.
(9) Do exactly what the task asks for, nothing more. Do not create, modify, or delete any data that the task does not explicitly require. If the task only asks you to read information, do not write or modify anything. If the task asks you to perform a specific action, perform only that action using the most direct approach available.
(10) If the task refers to another source of information (for example, a message, email, or request from someone), always read the actual content from that source first -- it may contain important details, constraints, or filters not stated in the task description.
(11) When a task depends on the current date, time, or day of the week, always determine it programmatically using the datetime python library -- never guess or assume.

Using these APIs, now generate code to solve the actual task:

My name is: {{ supervisor.first_name }} {{ supervisor.last_name }}. My personal email is {{ supervisor.email }} and phone number is {{ supervisor.phone_number }}.
Task: {{ instruction }}
\end{lstlisting}
\end{figure*}

\begin{figure*}[t]
\begin{lstlisting}[style=prettyprompt, captionpos=b, caption={OfficeBench agent prompt (1/2): the chat system message, including operating rules, the safety constraint, the \texttt{<think>}/\texttt{<answer>} response format, a one-shot example interaction, and the runtime app list injection point (\texttt{\{app\_introduction\}}).}, label={lst:officebench_prompt_system}]
Today is {date} ({weekday}). The current time is {time}. You are an AI assistant for user {username}.
You can help solve the task step by step.
You can interact with an operation system and use apps to solve the task.
You must follow the instructions and use the given json format to call APIs.
You can only generate one action at a time.
You can find files for your task in `{testbed_data_path}`. If you don't know the filenames, please switch to shell app and call commands to list the directory.
*Safety*: You may only manipulate/create files/folders under the `{testbed_data_path}` directory. You are not allowed to access any other data files or directories.
When responding, follow this thought-answer template and output the next step:
<think> [your reasoning here] </think>
<answer> [your JSON-formatted action here] </answer>
Strictly follow this format with no extra text. Below is an example interaction following this template.

##Task: Remind Bob that he has a dinner at 5/18/2024 7:00 p.m.
<think> To add a reminder for Bob's dinner, I need to create an event in his calendar. First, I must switch to the calendar app. </think>
<answer> {"app": "system", "action": "switch_app", "target_app": "calendar"} </answer>
Observation: "Successfully switched to app: calendar"
<think> Now that I'm in the calendar app, I should create an event for Bob's dinner at 7:00 p.m. on May 18, 2024. </think>
<answer> {"app": "calendar", "action": "create_event", "user": "Bob", "summary": "Dinner", "time_start": "2024-05-18 19:00:00", "time_end": "2024-05-18 20:00:00"} </answer>
Observation: "Successfully created a new event to Bob's calendar."
<think> The event has been successfully added to Bob's calendar. My task is complete. </think>
<answer> {"app": "system", "action": "finish_task", "answer": "None"} </answer>

You have following apps installed in the system:
{app_introduction}
\end{lstlisting}
\end{figure*}

\begin{figure*}[t]
\begin{lstlisting}[style=prettyprompt, captionpos=b, caption={OfficeBench agent prompt (2/2): the three per-step user prompt templates appended at each turn. The template is selected by whether an app is currently active and whether prior steps exist as history.}, label={lst:officebench_prompt_steps}]
=== prompt_undecided_app (no app currently active) ===
##Task: {task}
##Available apps: {available_apps}
##Instruction:
 - choose an app from the available apps: {"app": "system", "action": "switch_app", "target_app": [THE_APP_YOU_CHOOSE]}

=== prompt_undecided_app_w_history (no app active, prior steps exist) ===
##Task: {task}
##History:
{history}##Available apps: {available_apps}
##Instruction:
 - choose an app from the available apps: {"app": "system", "action": "switch_app", "target_app": [THE_APP_YOU_CHOOSE]}

=== prompt_decided_app (an app is currently active) ===
##Task: {task}
##Current app: {current_app}
##Instruction: Choose one action from the list as the next step.
{detailed_instruction} - switch to another app among {available_apps}: {"app": "system", "action": "switch_app", "target_app": [THE_APP_YOU_CHOOSE]}
 - finish the task with your answer as None if the task is not a question: <think> I'm finished the task. </think>
<answer> {"app": "system", "action": "finish_task", "answer": "None"} </answer>
 - finish the task with your answer if the task is a question: <think> I'm finished and the answer is [answer] </think>
<answer> {"app": "system", "action": "finish_task", "answer": [ANSWER]} </answer>
When responding, follow this thought-answer template and output the next step:
<think> [your reasoning here] </think>
<answer> [your JSON-formatted action here] </answer>
\end{lstlisting}
\end{figure*}

%% file: tables/judge_prompt.tex
%
%
%

\paragraph{AppWorld trajectory judge prompt.}
\Cref{lst:judge_appworld_system,lst:judge_appworld_user} show the prompts used by the LLM judge that filters AppWorld trajectories. The system prompt (\Cref{lst:judge_appworld_system}) defines nine requirements (documentation use, context understanding, strategy, information flow, parameter grounding, iteration, completion via \texttt{apis.supervisor.complete\_task}, deduplication, relevance) that a passing trajectory must satisfy. The user prompt (\Cref{lst:judge_appworld_user}) wraps the task and the agent's interaction trace in \texttt{<task>} / \texttt{<trajectory>} tags and asks for a YES/NO verdict on a line beginning with \texttt{`ANSWER:'}.

\begin{figure*}[t]
\begin{lstlisting}[style=prettyprompt, captionpos=b, caption={AppWorld trajectory judge -- system prompt. Defines nine requirements that a passing trajectory must satisfy.}, label={lst:judge_appworld_system}]
You are an AI model with strong reasoning and code understanding capabilities. You are provided with a user task within <task></task> XML tags and an agent's solution trajectory for this task within <trajectory></trajectory> XML tags. The provided trajectory consists of multiple steps, each containing agent's reasoning, code generated by the agent that includes one or more API calls and the corresponding code execution output.

Your job is to analyze whether the agent's solution trajectory correctly and completely solves the given task. DO NOT use your own factual world knowledge when verifying. The code execution output may contain synthetic data - focus on whether the agent's logic and approach reflects how the given task should be solved.

The solution MUST satisfy these requirements:

1. Always follows the documentation:
   - An API is never called without querying its documentation first.

2. Understands full context of the task:
   - If the task refers to prior activity, the agent should retrieve relevant data to fully understand the user's intent.

3. Follows correct strategy:
   - The agent breaks down the task into appropriate steps and executes them in logical order.

4. Proper information flow:
   - Retrieves necessary information through API calls
   - Uses outputs from earlier code in subsequent code (e.g., using returned access_token, IDs, etc.)
   - Properly chains API calls (login -> authenticated requests)

5. Grounded API parameters:
   - API input parameters either come from the task description, or the API descriptions, or the code from previous steps, or the previous code execution outputs, and must match the API documentation

6. Handles iteration correctly:
   - When the task requires searching or aggregating all possible entities of a particular type, the agent's code should iterate through all of them
   - Correctly implements loops or pagination with appropriate page limit value (based on the documentation) to access all data when required
   - Properly tracks state values across iterations (max value, matching items, etc.)
   - When an API returns a list, the agent should go through it to correctly identify the target item(s) instead of assuming that the target items are at the beginning of the list

7. Reaches completion:
   - If the task asks for some information (e.g., 'How many songs?', 'What is the email?'), the value of the answer provided to `apis.supervisor.complete_task(answer=...)` should answer the user's question correctly and completely based on the retrieved data.
   - If the task does not require an answer, `apis.supervisor.complete_task()` must be called without passing any argument value.

8. Avoids duplication:
   - The agent does not make redundant API calls that retrieve the same data multiple times unnecessarily.

9. Avoids irrelevant actions:
   - The agent does not perform actions that are completely irrelevant to solving the task.

Minor inefficiencies are acceptable (e.g., retrieving slightly more data than needed).
\end{lstlisting}
\end{figure*}

\begin{figure*}[t]
\begin{lstlisting}[style=prettyprompt, captionpos=b, caption={AppWorld trajectory judge -- user prompt. The agent's full interaction is inserted at \texttt{\{interaction\_trace\}}.}, label={lst:judge_appworld_user}]
Here is the task:
<task>
My name is {first_name} {last_name}. My personal email is {email} and phone number is {phone_number}.
{instruction}
</task>

Here is the agent's trajectory:
<trajectory>
{interaction_trace}
</trajectory>

Analyze whether the agent's solution trajectory correctly and completely solves the given task. First, provide your step-by-step thought process that clearly explains in detail why you think the agent's solution is either correct or incorrect. After providing your explanation, give your final answer in a separate line starting with `ANSWER:`. Your final answer must be either YES or NO. DO NOT provide anything else after the final YES/NO answer.
\end{lstlisting}
\end{figure*}

\paragraph{OfficeBench trajectory judge prompt.}
\Cref{lst:judge_officebench_system,lst:judge_officebench_user} show the prompts used by the LLM judge that filters OfficeBench trajectories. The OfficeBench system prompt (\Cref{lst:judge_officebench_system}) is rewritten from the AppWorld version to match OfficeBench's action format: actions are JSON \texttt{<answer>}-blobs rather than Python API calls, the agent must \texttt{switch\_app} before issuing app-specific actions, completion is signalled by \texttt{system.finish\_task} (not \texttt{apis.supervisor.complete\_task}), and many tasks are graded by side effects on files under \texttt{tasks/<task\_id>/testbed/data/}. The user prompt (\Cref{lst:judge_officebench_user}) follows the same \texttt{<task>}/\texttt{<trajectory>} structure but uses first-name only (OfficeBench supervisors expose nothing else).

\begin{figure*}[t]
\begin{lstlisting}[style=prettyprompt, captionpos=b, basicstyle=\ttfamily\scriptsize, caption={OfficeBench trajectory judge -- system prompt. Defines seven requirements tailored to OfficeBench's JSON-action interface and file-side-effect grading.}, label={lst:judge_officebench_system}]
You are an AI model with strong reasoning and code understanding capabilities. You are provided with a user task within <task></task> XML tags and an agent's solution trajectory for that task within <trajectory></trajectory> XML tags. The trajectory consists of multiple steps; each step contains the agent's JSON-formatted action wrapped as `<answer> {...} </answer>` and the observation returned by the OfficeBench environment.

OfficeBench is a multi-app office assistant: the agent can switch between apps  and the meta `system` app. To use an app's action the agent must first issue `{"app":"system","action":"switch_app","target_app":"<app>"}`. The agent terminates the task with `{"app":"system","action":"finish_task","answer":<value>}`, where `<value>` is the literal answer for question-style tasks or the string `"None"` for tasks that don't ask for an answer. Files for the task live under `tasks/<task_id>/testbed/data/` and the agent may only read/write under that directory.

Your job is to decide whether the agent's trajectory correctly and completely solves the given task. DO NOT use your own factual world knowledge for verification. Observation contents are synthetic — judge only from the logic and evidence the trajectory itself shows. In particular, do not penalize the agent because a filename or file content looks invented: whether a file exists is established by the observations, not by any real filesystem.

The solution MUST satisfy these requirements:

1. Reaches completion via finish_task:
   - The trajectory ends with a `system.finish_task` action and an observation indicating the task finished. A trajectory cut off by max-steps without `finish_task` does NOT pass.

2. Correct answer (for question-style tasks):
   - If the task asks a question (e.g., 'how many ...', 'what is the average ...'), the `answer` field of the final `finish_task` action must be the correct value, computable from values shown in earlier observations.
   - If the task does not ask for an answer, `answer` should be `"None"`.

3. Required side effects are produced:
   - When the task requires creating or converting a file (e.g., 'save it as a 'PDF', 'export the table to Excel', 'write the summary to a Word document'), the "agent must actually call the action that produces it and receive a success observation. Producing the intermediate artifact alone does NOT satisfy a task that asks for the converted output
   
   - Output Paths must be consistent with what the task asked for, and files consumed by a later step must be the ones written by an earlier step.

4. Information correctness and grounding:
   - Values written into files, emails, or events (names, scores, dates, titles) must come from the task description or from prior observations. The agent must not invent values that are supposed to be extracted.
   - When deriving data from a source (a spreadsheet, a note, a document), the agent must actually read it before writing derived result.

5. Correct app workflow:
   - Each app-specific action is executed while that app is active. A few wrong-app errors followed by `switch_app` recovery are acceptable; persistent thrashing without progress is not.

6. Action format:
   - Actions are valid JSON inside `<answer> ... </answer>`. Repeated format errors that prevent progress should fail the trajectory.

7. No unjustified shortcuts, fabrications, or irrelevant actions:
   - The agent does not skip required reads or conversions, write placeholder content where specific content was asked for, or perform actions unrelated to the task.
   -Using 'shell' to discover filenames is fine; using it to substitue for an app action the task requires is not.

Minor inefficiencies are acceptable (extra reads, an unnecessary app switch). What matters is that the required end-state - the right files, the right answer, the `finish_task` call - is achieved on evidence visible in the trajectory.
\end{lstlisting}
\end{figure*}

\begin{figure*}[t]
\begin{lstlisting}[style=prettyprompt, captionpos=b, caption={OfficeBench trajectory judge -- user prompt. OfficeBench supervisors expose only \texttt{first\_name}, so the task header is shorter than AppWorld's.}, label={lst:judge_officebench_user}]
Here is the task:
<task>
My name is {first_name}.
{instruction}
</task>

Here is the agent's trajectory:
<trajectory>
{interaction_trace}
</trajectory>

Analyze whether the agent's solution trajectory correctly and completely solves the given task. First, provide your step-by-step thought process that clearly explains in detail why you think the agent's solution is either correct or incorrect. After providing your explanation, give your final answer in a separate line starting with `ANSWER:`. Your final answer must be either YES or NO. DO NOT provide anything else after the final YES/NO answer.
\end{lstlisting}
\end{figure*}

%% file: sections/7_synthetic_apps.tex
\section{Synthetic Application Suite}
\label{sec:synthetic_apps}

We construct a suite of 65 synthetic apps spanning twelve domains. Each application is modeled as a set of REST APIs, fully specified with typed endpoints, parameter constraints, and structured response schemas. The suite comprises 1,062 API endpoints across the 65 apps.

\subsection{Application Generation}

The 12 application domains were manually curated according to two principles: \emph{real-world fidelity}, whereby each domain corresponds to a widely used category of software, and \emph{functional diversity}, whereby the collection as a whole requires agents to exercise qualitatively different interaction patterns, data models, and reasoning skills.

We manually defined a single example application, consisting of an application name, a short description, and a list of key agent-facing functionalities (see \Cref{lst:seed_domain}). Using this entry as a structural in-context example, we prompted Gemini-2.5-Pro to generate diverse applications for each of the 12 target domains.

\begin{figure*}[ht!]
\begin{lstlisting}[language=json,
  basicstyle=\ttfamily\scriptsize,
  breaklines=true,
  style=prettyprompt,
  columns=fullflexible,
  captionpos=b,
  caption={A representative entry from the manually curated  application used to prompt Gemini for further application expansion. Each entry specifies the \texttt{general\_domain}, the specific \texttt{name}, a \texttt{description}, and a comma-separated list of agent \texttt{functionalities}.},
  label={lst:seed_domain}]
{
  "general_domain": "Finance & Payments",
  "name": "Peer-to-Peer Money Transfer",
  "description": "A digital wallet app that lets users send and receive
      money, split bills, and manage payment methods.",
  "functionalities": "send money, request payments, view transaction
      history, manage payment cards, add/withdraw balance,
      like and comment on transactions"
}
\end{lstlisting}
\end{figure*}


\Cref{tab:domains} summarizes the twelve domains and their constituent app counts for the 65 synthetic apps.

\begin{table*}[t]
\centering
\small
\setlength{\tabcolsep}{12pt}
\begin{tabular}{lcp{8.5cm}}
\toprule
\textbf{Domain} & \textbf{\#Apps} & \textbf{Representative interaction patterns} \\
\midrule
Document Authoring \& Conv.  & 4  & Author slides, fill forms and templates, translate and convert documents \\
Files \& Data Operations     & 6  & List and move files, edit spreadsheets, query databases, plot charts, archive directories \\
Social \& Communication      & 7  & Post, follow, message, react, moderate content \\
E-Commerce \& Marketplace    & 5  & Browse, bid, purchase, review, manage listings \\
Productivity \& Proj.\ Mgmt  & 11 & Create tasks, manage boards, schedule events, sign docs, take notes, look up contacts \\
Entertainment \& Media       & 7  & Stream, subscribe, rate, bookmark, manage playlists \\
Travel \& Transportation     & 5  & Book, track, navigate, estimate fares, manage trips \\
Food \& Health               & 5  & Order food, log meals, track fitness, manage health records \\
Developer \& Business Tools  & 5  & Manage repos, file tickets, run campaigns, track leads \\
Finance \& Payments          & 4  & Transfer funds, trade assets, manage accounts, pay bills \\
Learning \& Hobbies          & 3  & Complete lessons, track workouts, check weather \\
Local \& Community           & 3  & Discover venues, RSVP to events, join groups \\
\midrule
\textbf{Total}               & \textbf{65} & \\
\bottomrule
\end{tabular}
\caption{The twelve functional domains of the synthetic app suite, obtained by manual inspection of real-world software categories. Each domain was defined to represent a distinct interaction paradigm for LLM API-calling agents.}
\label{tab:domains}
\end{table*}

\subsection{API Schema Design}

For each application, we generated a complete REST API schema by prompting Gemini-2.5-Pro with (1) the target application description drawn from our domain taxonomy, and (2) six representative $\langle\mathtt{banking\_app}\rangle$ APIs from AppWorld as structural few-shot examples. The six examples were deliberately selected to cover the full range of structural patterns that appear in the suite: a public authentication API (\texttt{login}), a simple authenticated \texttt{GET} (\texttt{show\_account}), a resource creation \texttt{POST} with required and optional parameters (\texttt{create\_transaction}), a complex list API with filters, pagination, and \texttt{sort\_by} (\texttt{show\_transactions}), a stateful action on a sub-resource (\texttt{like\_transaction}), and a sub-resource \texttt{GET} by ID (\texttt{show\_payment\_card}). \Cref{lst:example_apis} shows two of these six examples.

\begin{figure*}[t]
\begin{lstlisting}[language=json,
  style=prettyprompt,
  basicstyle=\ttfamily\scriptsize,
  breaklines=true,
  breakatwhitespace=false,
  columns=fullflexible,
  captionpos=b,
  caption={Two of the six $\langle\mathtt{banking\_app}\rangle$ APIs provided as few-shot structural examples to Gemini: a complex list endpoint with filters and pagination (top) and a stateful sub-resource action (bottom). These two examples illustrate the range of structural patterns Gemini is expected to replicate.},
  label={lst:example_apis}]
{
  "app_name": "$\langle\mathtt{banking\_app}\rangle$",  "api_name": "show_transactions",
  "path": "/$\langle\mathtt{banking\_app}\rangle$/transactions",  "method": "GET",
  "description": "Search or show a list of your transactions.",
  "parameters": [
    {"name": "access_token", "type": "string",  "required": true,
     "default": null, "constraints": []},
    {"name": "query",        "type": "string",  "required": false,
     "default": "", "constraints": []},
    {"name": "min_amount",   "type": "number",  "required": false,
     "default": 0.01, "constraints": ["value > 0.0"]},
    {"name": "direction",    "type": "string",  "required": false,
     "default": null,
     "constraints": ["value in ['sent', 'received']"]},
    {"name": "page_index",   "type": "integer", "required": false,
     "default": 0,    "constraints": ["value >= 0"]},
    {"name": "page_limit",   "type": "integer", "required": false,
     "default": 5,    "constraints": ["value >= 1, <= 20"]},
    {"name": "sort_by",      "type": "string",  "required": false,
     "default": null, "constraints": []}
  ],
  "response_schemas": {
    "success": [{"transaction_id": 1, "amount": 0.0,
      "description": "string", "created_at": "2019-01-01T00:00:00",
      "private": true,
      "sender":   {"name": "string", "email": "user@example.com"},
      "receiver": {"name": "string", "email": "user@example.com"}}],
    "failure": {"message": "string"}
  }
},
{
  "app_name": "$\langle\mathtt{banking\_app}\rangle$",  "api_name": "like_transaction",
  "path": "/$\langle\mathtt{banking\_app}\rangle$/transactions/{transaction_id}/likes",
  "method": "POST",
  "description": "Like a transaction.",
  "parameters": [
    {"name": "transaction_id", "type": "integer", "required": true,
     "default": null, "constraints": []},
    {"name": "access_token",   "type": "string",  "required": true,
     "default": null, "constraints": []}
  ],
  "response_schemas": {
    "success": {"message": "string"},
    "failure": {"message": "string"}
  }
}
\end{lstlisting}
\end{figure*}

\paragraph{Prompt structure.}
The full Gemini prompt is structured in five parts, shown in abbreviated form in \Cref{lst:gemini_prompt}. The domain name and description from the domain taxonomy (\Cref{lst:seed_domain}) are injected at the top of the prompt as a \emph{target header} (Part~1), which tells Gemini which specific domain to generate and which key workflows to cover. Part~(2) specifies the required \texttt{app\_description} and the API coverage requirements (core CRUD, sub-resources, actions, search/filter, social features, notifications, payments, file downloads). Part~(3) defines the exact JSON output schema for each API object and each parameter object. Part~(4) states mandatory conventions: the \texttt{access\_token} parameter ordering rule, the 10 standard auth/account APIs every app must include, pagination conventions, and response schema placeholder formats. Part~(5) provides the six $\langle\mathtt{banking\_app}\rangle$ example APIs. A diversity constraint is appended when generating multiple apps in sequence, listing already-generated app names that must not be duplicated.

\begin{figure*}[t]
\begin{lstlisting}[style=prettyprompt,
  basicstyle=\ttfamily\scriptsize,
  breaklines=true,
  columns=fullflexible,
  captionpos=b,
  caption={Abbreviated structure of the Gemini prompt used for synthetic app generation. Domain entries are loaded from \texttt{domains\_subdomains.json} at runtime; \texttt{<domain.*>} placeholders are filled from the current entry. Ellipses (\texttt{...}) denote sections paraphrased for brevity.},
  label={lst:gemini_prompt}]
# Load domain entries from file
domains = json.load(open("domains_subdomains.json"))

# For each domain entry, build and send this prompt:
# --- Part 1: Target header (domain entry injected from domains_subdomains.json) ---
Your job is to invent ONE new <domain.name> app
(in the <domain.general_domain> category).

App context: <domain.description>
Key workflows to cover: <domain.functionalities>

IMPORTANT: Invent a new, distinct app name.
           Do not reuse any well-known brand name.

# --- Part 2: Coverage requirements ---
You are an expert API designer building a new mobile app backend
from scratch. 

Beyond the 10 standard account management APIs, cover:
  1. Core entities: CRUD for each major resource
  2. Sub-resources: nested resources (e.g. order items, comments)
  3. Actions: stateful transitions (e.g. cancel_order, approve_request)
  4. Search & filtering: list APIs with filters, date ranges, enums
  5. Social features: follows, likes, ratings, reviews, comments
  6. Notifications: show_notifications, mark_notification, delete_notification
  7. Payments: payment methods, charges, refunds, receipts
  8. File downloads: receipts/invoices following $\langle\mathtt{banking\_app}\rangle$ download_*_receipt

# --- Part 3: Output format ---
A JSON object { "<app_name>": { "app_description": "...", "apis": [...] }}
Each API: { "app_name", "api_name", "path", "method",
  "description", "parameters", "response_schemas" }
Each parameter: { "name", "type", "required", "description",
  "default", "constraints" }

# --- Part 4: Conventions ---
access_token: required on all non-public APIs; order is
  required params -> access_token -> optional params.
Public APIs (no access_token): login, signup,
  send_verification_code, send_password_reset_code,
  reset_password, verify_account.
10 required standard APIs: signup, login, logout, show_account,
  update_account, send_verification_code, verify_account,
  send_password_reset_code, reset_password, delete_account.
Pagination on all list APIs: page_index (default 0),
  page_limit (default 5), sort_by (optional).
Response placeholders: "string", 1, 0.0, true,
  "2019-01-01T00:00:00", "user@example.com".
Constraints: "length >= 1", "value >= 0",
  "value in ['a','b']", "value is email address".

# --- Part 5: Few-shot examples ---
## Example: 6 representative $\langle\mathtt{banking\_app}\rangle$ APIs
[... six full API objects as shown in \Cref{lst:example_apis} ...]

# --- Diversity constraint (appended when N > 1) ---
The following apps have already been generated. Your new app MUST
be from a completely different domain: `<app1>`, `<app2>`, ...

Output only the JSON object, no explanation, no markdown.
\end{lstlisting}
\end{figure*}

\paragraph{Verification.}
After each Gemini response, a rule-based verifier checks structural invariants before accepting the schema: all required top-level keys are present; all 10 standard APIs are included; each API has the seven required fields; paths start with \texttt{/<app\_name>/}; HTTP methods are valid; \texttt{access\_token} is absent on public APIs and present on all others; \texttt{access\_token} does not appear after optional parameters; \texttt{response\_schemas} has both \texttt{success} and \texttt{failure} keys; all parameter types are from the allowed set; and at least one domain-specific API exists beyond the standard 10. Schemas that fail verification are discarded and regenerated up to a configurable retry limit.

\paragraph{Schema structure.}
Each API is defined by eight fields: \texttt{app\_name}, \texttt{api\_name}, \texttt{path}, \texttt{method} (\texttt{GET} / \texttt{POST}/ \texttt{PATCH} / \texttt{DELETE}), \texttt{description}, \texttt{parameters}, \texttt{response\_schemas}, and \texttt{canary\_string}. Specifications are authored using four Python helpers that enforce structural uniformity across all applications: \texttt{api\_()} constructs a full API entry; \texttt{p()} constructs a typed parameter; \texttt{tok()} is shorthand for the standard \texttt{access\_token} parameter; and \texttt{paged()} appends the three standard pagination parameters to a parameter list.

\paragraph{Design conventions.}
We enforced the following conventions across all 65 applications to ensure a consistent interface vocabulary:

\begin{itemize}[leftmargin=1.5em, itemsep=2pt]
    \item \textbf{Authentication.} Every application exposes a \texttt{login} API returning an \texttt{access\_token}, which is required on all authenticated APIs, mirroring the pattern in the original AppWorld applications.
    \item \textbf{Pagination.} All list-returning APIs accept \texttt{page\_index} (0-based, default 0), \texttt{page\_limit} (default 5), and \texttt{sort\_by} (attribute prefixed with \texttt{+}/\texttt{-} for ascending/descending order).
    \item \textbf{Type-safe response schemas.} Response schemas use typed placeholders: \texttt{``string''} for text fields, \texttt{1} for integer IDs, \texttt{0.0} for floats, \texttt{``2019-01-01T00:00:00''} for timestamps, and \texttt{``user@example.com''} for email fields, never domain-specific invented values.
    \item \textbf{Constraint notation.} Parameter constraints follow a strict vocabulary: \texttt{``length >= 1''}, \texttt{``value >= 0.0, <= 5.0''}, \texttt{``value in ['a', 'b']''}, and \texttt{``value is email address''}. No other formats are valid.
    \item \textbf{Nested sub-resources.} Where a resource logically contains an embedded sub-resource (\eg, a ride has a driver; a booking has a host), the response schema includes a nested object rather than flattening all fields.
    \item \textbf{Failure responses.} The \texttt{failure} branch of every API is uniformly \texttt{\{``message'': ``string''\}} across all 65 applications.
\end{itemize}

\subsection{Schema Conversion}

The Python specification files are serialized to a consolidated JSON file and then converted per app into AppWorld's canonical format using Gemini-3.1-Pro. For each app, a structured prompt is constructed containing: (1) reference API specifications as few-shot demonstrations; (2) explicit formatting rules for description style, constraint vocabulary, and response schema placeholder conventions; and (3) the full API specification for the target application. Gemini returns a JSON object.

Each API entry is annotated with a \emph{canary string}, where the MD5 prefix is the first seven hex characters of the application name's hash and the UUID is freshly generated per API. Canary strings serve two purposes: they enable traceability of downstream generated data back to its source application, and they provide a signal for detecting schema memorization during model evaluation. The conversion pipeline includes automatic retry on JSON parse failures and falls back to a deterministic structural converter for any applications where LLM conversion fails, ensuring all 65 applications produce valid output.